\documentclass[lettersize,journal]{IEEEtran}
\usepackage{amsmath,amsfonts}
\usepackage{algorithm}
\usepackage{array}
\usepackage{textcomp}
\usepackage{stfloats}
\usepackage{url}
\usepackage{verbatim}
\usepackage{graphicx}
\usepackage{cite}
\usepackage{hyperref}       
\usepackage{url}            
\usepackage{booktabs}       
\usepackage{amsfonts}       
\usepackage{nicefrac}       
\usepackage{microtype}      
\usepackage{xcolor}         
\usepackage{graphicx}
\usepackage{caption}
\usepackage{multirow}
\usepackage{subfigure}
\usepackage{amsmath}
\usepackage{algorithm}
\usepackage{algpseudocode}
\usepackage{amsmath}
\usepackage{amsmath,lipsum}
\usepackage{cuted}
\begin{document}

\title{Linear Leaky-Integrate-and-Fire Neuron Model Based Spiking Neural Networks and Its Mapping Relationship to Deep Neural Networks}

\author{Sijia Lu,~\IEEEmembership{Student Member, IEEE}, Feng Xu, ~\IEEEmembership{Senior Member, IEEE}

\thanks{This paper was produced by the IEEE Publication Technology Group. They are in Piscataway, NJ.}
\thanks{Manuscript received April 19, 2021; revised August 16, 2021.}}

\markboth{Journal of \LaTeX\ Class Files,~Vol.~14, No.~8, August~2021}%
{Shell \MakeLowercase{\textit{et al.}}: A Sample Article Using IEEEtran.cls for IEEE Journals}


\maketitle

	\begin{abstract}

Spiking neural networks (SNNs) are brain-inspired machine learning algorithms with merits such as biological plausibility and unsupervised learning capability. Previous works have shown that converting Artificial Neural Networks (ANNs) into SNNs is a practical and efficient approach for implementing an SNN. However, the basic principle and theoretical groundwork are lacking for training a non-accuracy-loss SNN. This paper establishes a precise mathematical mapping between the biological parameters of the Linear Leaky-Integrate-and-Fire model (LIF)/SNNs and the parameters of ReLU-AN/Deep Neural Networks (DNNs). Such mapping relationship is analytically proven under certain conditions and demonstrated by simulation and real data experiments. It can serve as the theoretical basis for the potential combination of the respective merits of the two categories of neural networks.

\end{abstract}


\section{Introduction}

In recent decades, Artificial Intelligence (AI) has taken a path that has been rising, then falling, and is now under steady development. Based on the understanding of the human cerebral cortex's mechanism, ANN is formulated and becomes one of the primary directions for AI, called connectionism \cite{mcculloch1943logical}. ANNs are composed of artificial neurons (ANs) connected as a graph. The weights of the connections, mimicking the cerebral cortex's synapses, represent the network's plasticity and can be trained via gradient descent \cite{ruder2016overview} in supervised learning tasks. With a large amount of annotated training data, a deep large-scale network structure, and computing power, DNNs have achieved great successes in many application fields. They have become the most popular AI technology. The performance of a DNN can reach the human level on specific tasks, such as image recognition \cite{krizhevsky2012imagenet,he2016deep,8627998,7927455}, instance segmentation \cite{cao2020sipmask}, speech understanding \cite{hinton2012deep}, strategic game playing \cite{mnih2013playing}, etc.

DNNs employ a hierarchical structure with an exponentially-growing representation capacity. Such deep network structure was studied as early as the 1980s, but it was found difficult to train due to the vanishing of backpropagated gradients \cite{articleIvakhnenko,articleIvakhnenko1971,SCHMIDHUBER201585}. This problem was not solved until the deep learning era when the much simpler activation function called Rectified Linear Unit (ReLU) was used instead of conventional nonlinear functions such as the sigmoid \cite{jarrett2009best,articleGlorot,Choromanska2014The}. Equipped with the ReLU activation function, DNNs have gained a powerful fitting capability on large-scale complex data. DNN is considered second-generation neural networks \cite{maass1997networks}. It is widely considered that DNN’s great success is attributable to big data, powerful computational technology (such as GPU), and training algorithms. 

As DNNs are widely applied in real applications, limitations are becoming apparent. For example, strong dependence on labeled data and non-interpretability are considered drawbacks of deep learning. With the increase of layers and parameters, DNNs require many annotated data and computing power for training. However, current research mainly focuses on network architecture and algorithms designed for specific AI tasks. A technical approach to general artificial intelligence aims to break the limitations that remain studied. In this regard, many methods have been proposed, including SNN \cite{maass1997networks}, which is regarded as the third generation of neural networks. SNN uses spiking neurons primarily of the leaky-Integrate-and-Fire (LIF) type \cite{lapicque1907recherches}, which exchange information via spikes. Due to its accurate modeling of biological neural network dynamics, SNN is the most popular brain-inspired AI approach \cite{tan2020spiking}.  There have been extensive studies of SNN-derived neural networks, such as full connected SNN \cite{diehl2015unsupervised}, deep SNN \cite{ILLING201990,tavanaei2019deep} and convolution SNN \cite{KHERADPISHEH201856}. The learning mechanism of SNN includes supervised learning (such as spike backward propagation) \cite{KULKARNI2018118,WANG2020258}, unsupervised learning (such as spiking timing-dependent plasticity) \cite{Tavanaei2016Bio,Nazari2018Spiking}, and reinforcement learning \cite{8356226}. 

However, SNNs have not yet achieved the performance of DNN in many tasks. One of the most effective training algorithms is to transfer the trained weights of DNNs to SNNs with the same structure \cite{cao2015spiking,kim2020spiking,sengupta2019going}. Establishing an effective SNN training algorithm or transformation mechanism is a challenging task. The fundamental question on the relationship between the second and third-generation neural networks is unclear.

The major contributions of this paper are as follows:

\begin{itemize}
	\item The parameter mapping relationship between the Linear LIF neuron model and the ReLU-AN model is established.
	\item Inspired by the perspective of biology as well as the proposed equivalence, ReLU activation function is proved to be the bridge between SNNs and DNNs.
	\item Experiments conducted on the MNIST and the CIFAR-10 datasets demonstrate the effectiveness and superiority of the proposed SNN composed of the linear LIF model. The experimental validation under various simulation conditions are presented to prove the equivalence.
\end{itemize}

The rest of the paper is organized as follows. Section \ref{motivation} explains the motivation of this study. Section 3 summarizes the related studies on ReLU-AN and the LIF model. Section 4 defines equivalence and presents the mapping relationship between the LIF model and ReLU-AN model. Simulations and analyses from single neuron to deep neural networks are carried out in Section 5. Finally, we make a brief conclusion and state the future opportunities in Section 6.


\section{Motivation}\label{motivation}

\subsection{Bridge the gap between ANN and SNN}

Brain science and cognitive neuroscience have been one of the essential sources of inspiration for artificial intelligence \cite{bear2007neuroscience,10.3389/fncom.2016.00094}. From this perspective, we want to establish the relationship between ANN and SNN, which may bridge artificial intelligence and computational neuroscience. We believe that ANN, the most powerful AI  in real applications, and SNN, the most biologically plausible technology, can learn from each other. 

The biological neural model's complex dynamics and non-differentiable operations make SNN lack scalable training algorithms. In this paper, we focus on the mechanism of transferring trained weights of DNN into SNN. While this method has achieved good results in target classification tasks, it has relatively strict limitations on pre-trained DNN, especially bias transformation. In the SNN conversion toolbox (SNN-TB) \cite{rueckauer2017conversion}, the bias is represented as a constant input current or an external spike input of constant rate proportional. However, we believe that bias in neuron model can be reflected in the biological neuron model, which we will show in the following simulations. In addition, the thickness and length of the axon of a neuron are different, and the neuron model parameters should also be different. This is not reflected in SNNs while some DNN-to-SNN algorithms use dynamic spiking threshold. We intend to establish the equivalent relationship between spiking neurons and artificial neurons and then the transformation mechanism of ANN.

\subsection{A biological explanation of ReLU}
DNNs use many layers of nested nonlinearity to fit massive amounts of data and perform better in machine learning tasks with ReLU.  We focus on the nonlinearity and sparsity of ReLU, but we do not have a deep understanding of why the ReLU performs better than the other activation function. Glorot \cite{articleGlorot} indicates that ReLU can bridge the gap between the computational neuroscience model and the machine learning neural network model. But under what conditions, i.e. coding algorithms and parameters, RELU can be equivalent to the biological model, and what is the mathematical mapping relationship between the two models. This is still a fundamental question in biologically inspired AI that remains unanswered.

\subsection{A new approach of unsupervised learning}

The unsupervised learning mechanism employed in SNN has a good biological basis and emphasizes the causal relations between the signals, which complements conventional machine learning. Unsupervised learning is generally regarded as a representation learning that estimates a model representing the distribution for a new input $x_n$ given previous inputs $x_{1}, x_{2}, \ldots, x_{n-1}$, expressed as $\mathrm{P}\left(x_{n} \mid x_{1}, x_{2}, \ldots, x_{n-1}\right)$ \cite{ghahramani2003unsupervised}. Computational neuroscience has provided a new idea for unsupervised learning mechanisms. Spiking Time Dependent Plasticity (STDP) \cite{abbott2000synaptic,Song2000Competitive,doi:10.1146/annurev.neuro.31.060407.125639,FALEZ2019418,TAVANAEI2018294}, is a temporally asymmetric form of Hebbian learning and is the most widely used unsupervised learning mechanism in SNNs. In the temporal dimension, the relation between the presynaptic action potential and the postsynaptic action potential regulates the neurons' weights, which is a feature unique to SNNs. Suppose we want to migrate such a natural learning mechanism in the time domain from SNNs to DNNs. In that case, we first need to establish a mathematical mapping relationship between SNN's neuron model and DNN’s neuron model.

\subsection{Inspire the development of artificial intelligence }	

SNN has its unique advantages in information transmission and learning mechanisms. Although ANN is historically brain-inspired, ANN and SNN are entirely different. First, SNN uses event-driven characteristics to reduce power consumption. SNN transfer and process the information via spike train \cite{tavanaei2019deep}, while DNN uses scalars to represent the neural signals. For the same task, e.g., image and voice recognition, the human brain typically consumes 10–20 watts \cite{articleJeong2016}, compared to hundreds of thousands of watts for DNNs running on a computer. Secondly, neuromimetic calculation is not a conventional von Neumann architecture but adopts an integrated structure of storage and calculation, storing information in neurons. The mechanism of time-domain processing in spike trains and Hebbian learning-based synaptic plasticity are considered potential routes to a more advanced artificial intelligence \cite{hebb1949organization,Song2000Competitive,denham2001dynamics}.

\section{Related work}

\subsection{ReLU Artificial Neuron}

An Artificial Neuron (AN) is a mathematical function that can model a biological neuron. McCulloch and Pitts proposed the Artificial Neuron model in 1943. It is also known as the M-P model and is still used today. As the basic unit of the neural network, it receives input signals from previous layer units or perhaps from an external source. Each input has an associated weight $\omega$, which can be adjusted to model the synaptic plasticity. Through an activation function $f( \ )$, the unit converts the integrated signal, i.e., the weighted sum of all the inputs, to obtain its output

\begin{equation}\label{ReLU}
	y_{i}=f(\sum_{j} \omega_{i j} x_{j} +b_{i})
\end{equation}
Here, $\omega_{i j}$ is the weight from unit $j$ to unit $i$, $b_i$ is the bias of unit $i$, and $f( \ )$ is the activation function. For the M-P model, the form of activation function is the Heaviside step function. The working model of neurons has two states, activation (1) and inhibition (0). The main idea of deep learning is still very similar to the perceptron proposed by Frank Rosenblatt many years ago, but the binary Heaviside step function is no longer used. Neural networks mostly use the ReLU activation function.

\begin{figure}[htb!]
	\centering
	\includegraphics[width=0.5\linewidth]{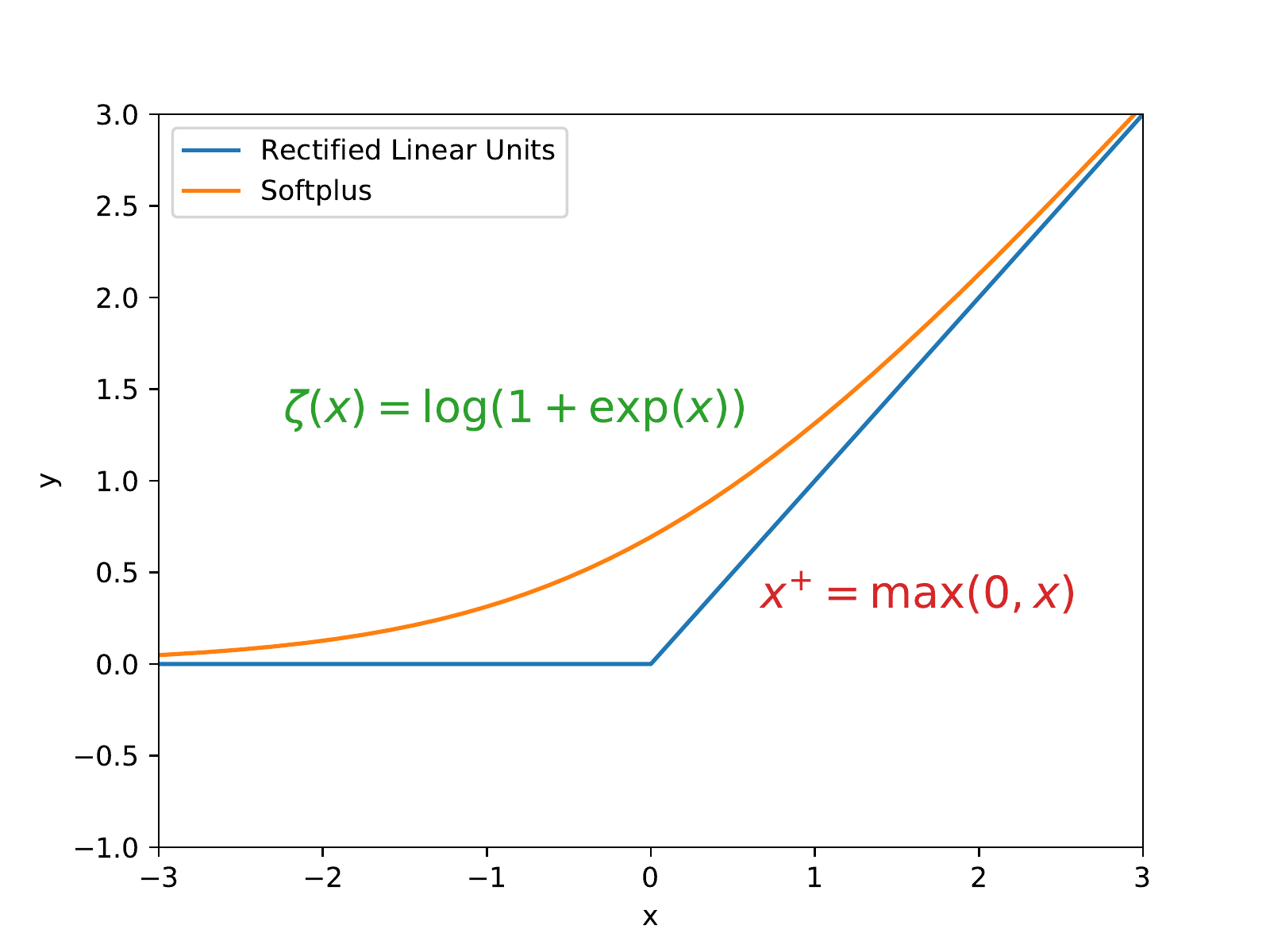}
	\caption{Rectified Linear Unit and softplus activation functions.}
	\label{fig:relu}
\end{figure}

ReLU has been developed for a long time. Cognitron \cite{fukushima1975cognitron} is considered the first artificial neural network using a multi-layered and hierarchical design. This paper also proposed the initial form of ReLU, i.e., $max(0,x)$, shown in Fig.\ref{fig:relu}. The activation function, a rectification nonlinearity theory, applies to the Symmetric Threshold-Liner network dynamics \cite{HahnloserPermitted}. Pinto and Cox proposed a V1-like recognition system, in which the outputs of the Gabor filter will pass through a standard output nonlinearity---a threshold and response saturation \cite{pinto2008real}. Jarrett and Kavukcuoglu have proved that rectified nonlinearities are the single most crucial ingredient for deep learning \cite{jarrett2009best}. ReLU was then introduced to enhance the ability in the feature learning of restricted Boltzmann machines. Compared with the sigmoid function, it has achieved better image classification accuracy \cite{nair2010rectified}. Glorot (2011) showed that neurons with ReLU activation function have a better performance than hyperbolic tangent networks and analyzed the advantage of sparsity \cite{articleGlorot}. In the deep learning era, ReLU was crucial to the training of deep neural networks. 

The critical characteristics of ReLU are: A. Nonlinearity: Introducing nonlinearity is critical for deep neural networks. With the simple rectification, it provides the fundamental nonlinearity required for data fitting. B. Sparsity: Nearly half of the neuron's outputs are suppressed. This mechanism is similar to lateral inhibition in biological neural networks \cite{amari1977dynamics} and increases the neural network's sparsity.

\subsection{LIF Neuron model}

Here we present the development of biological neurons and analysis of LIF neuron's dynamic properties.

\subsubsection{Development of the Biological Neuron}

Research on biological neuron models can be dated back to the 1900s, referred to Lapicque model \cite{lapicque1907recherches,abbott1999lapicque}, which is employed in the calculation of firing times. The Lapicque model is considered the earliest form of the `integrate-and-fire model,' and it becomes the LIF model after adding an attenuation term. The LIF model is one of the most popularly used models for analyzing the nervous system's behavior \cite{diehl2015unsupervised,KHERADPISHEH201856}. Unlike the neuron models used for computing, some neuron models have also been created and applied to simulate real neuron propagation potentials. The Hodgkin--Huxley model (HH model) \cite{hodgkin1952quantitative} was proposed by analyzing the electric current flow through the surface membrane. We call it a simulation-oriented neuron model. However, it is not practically applied in general neural networks due to its computational complexity. For computational feasibility, simplified models have emerged, referred to as computation-oriented models. Izhikevich model \cite{izhikevich2003simple} is a simplification of the HH model based on the theory of dynamic systems. In this paper, we focus on the dynamic properties of the LIF model and analyze its equivalence with ReLU.

\subsubsection{The dynamic of LIF neuron model}



The LIF model can be modeled as a circuit composed of a resistor and a capacitor in parallel, which respectively represent the leakage and capacitance of the membrane \cite{articleTuckwell}. The integrate-and-fire  neuron model is described by the dynamics of the neuron's membrane potential (MP), $V(t)$,

\begin{equation}\label{lifbaseformula}
	C_m \frac{dV(t)}{dt}+\frac{V(t) - V_0}{R_m}=I_{inj},
\end{equation}
where $C_m$ and $R_m$ denote, respectively, the membrane capacitance and resistance, $V$	is the membrane potential of LIF model, $V_0$ is the resting potential, and $I_{inj}$ is the current injected into the neuron. The driving current can be split into two components, $I(t) = I_{C_m} + I_{R_m}$ \cite{gerstner2002spiking}. The first part on the left of Eq. \ref{lifbaseformula} represents the current passing through the capacitor during charging. According to the definition of capacitor $C = Q / U$ (where $Q$ is the charge and $U$ is the voltage), we find that the membrane capacitance current $I_{C_m} = C_m dV /dt$. The second part represents the leak of the membrane through the linear resistor $R_m$, and the membrane time constant  $\tau_m = C_m R_m$ of the `leaky integrator' is introduced \cite{burkitt2006a}.

Given the initial value of membrane potential and injected current $I_{inj}$, we can use the method of integrating factors \cite{maday1990operator} to solve the differential equations which defines the change of membrane potential, and the simplified equation can be expressed as

\begin{equation}\label{nernstplanckequation}
	V\left(t\right)=e^{-\frac{t-t_0}{\tau_m}}\left[\int_{t_0}^{t}{\frac{I_{inj}\left(t^\prime\right)}{C_m}e^\frac{t^\prime-t_0}{\tau_m}dt^\prime}+V\left(t_0\right)\right]
\end{equation}
where $I_{inj}$ is the injected current, $V(t_0)$ is the membrane potential at the initial time $t_0$, and we take the reset potential to be $V_{reset} = 0$ for the sake of simplicity. When the neuron has no input, i.e. $I_{inj} = 0$, the integral term in Eq.\ref{nernstplanckequation} is 0, and the membrane potential  decays exponentially on the basis of the initial potential $V(t_0)$; when there is input, the input current is integrated into the post-neuronal membrane potential. 

To explore a self-consistent neuron model, neurons' input and output forms should be the same.  The input current, $I_{inj}(t)$, is defined as the weighted summation of pre-synaptic spikes at each time step,
\begin{equation}
	I_{inj}(t)=\sum_{i=1}^{n^l}\omega_{i} \cdot \sum_{j = 1}^n{\delta(t- t_j)}
\end{equation}
where  $n^l$ indicates the number of pre-synaptic weights, $n$ is the number of spikes of pre-synaptic spike train, $\omega_{i}$ gives the connection weights between the pre-synaptic neuron $i$ and post-synaptic neuron, and $\delta$ is the Dirac function. The top of Fig. \ref{fig:continuouslif} shows the dynamics of membrane potentials of an LIF neuron with multiple input spikes.

Assuming that a spiking input signal with a period of $T$ (frequency $f =1/T$ and $t_j = j/f$), the value of MP can be described as:

\begin{equation}\label{CLIF}
	V(nT^+)=\frac{\omega}{C} \cdot \frac{1-e^{-nT/\tau_m}}{1-e^{T/\tau_m}}
\end{equation}

The membrane potential accumulates with the presence of inputs $ I(t) $. Once the membrane potential $V(t)$ exceeds the spiking threshold $ V_{th} $, the neuron fires an action potential, and the membrane potential $V(t)$ goes back to the resting potential $ V_0 $. The LIF model is a typical nonlinear system. Three discrete equations can describe the charge, discharge, and fire of the LIF model:

\begin{equation}
	\begin{aligned}
		H(t) & = f(V(t-1), I(t)) \\
		S(t) & = \Theta(H(t) - V_{th})
	\end{aligned}
\end{equation}
where the $H(t)$ is the membrane potential before spike, $S(t)$ is the spike train and $f(V(t-1), I(t))$ is the update equation of membrane potential.


\section{The mapping relationship between LIF neuron model and ReLU-AN model}


Converting CNNs into SNNs is an effective training method that enables mapping CNNs to spike-based hardware architectures. Many scholars believe that the theoretical equivalence between the spiking neuron model and the artificial neuron model is the basis of the transformation method. This section presents a mapping relationship between the "linear reset LIF model" and the ReLU-AN model.

Many differences exist between the neural network models used in machine learning and those used in computational neuroscience. \cite{articleGlorot} shows that the ReLU activation function can bridge the gap between these two neuron models, including the sparse information coding and non -linear. Mainly based on changing the activation function from tanh() to HalfRect(x) = max(x,0), which is named ReLU and is nowadays the standard model for the neuron in DNNs, \cite{cao2015spiking} proposed a method for converting trained CNN to SNN with slight performance loss. However, the theoretical groundwork of converting basic principles is lacking, and related research only shows the similarity between the LIF neuron model and the AN model. \cite{rueckauer2017conversion} present a one-to-one correspondence between an ANN unit and an SNN neuron and an analytical explanation for the approximation. On this basis, this paper further presents an exact correspondence between the parameters of the LIF neuron model and the ReLU-AN model.

\subsection{Linear leaky-integrate-and-fire model}

Once the membrane potential reaches the spiking threshold, an action potential will be exceeded. Then the membrane potential will be reset: 'Reset-to-Zero,' used, e.g., in \cite{diehl2015unsupervised}, reset the membrane potential to zero. 'Linear Reset' retains the attenuation term that exceeds the threshold:

\begin{equation}
V(t)=\left\{\begin{array}{cl}
H(t) \cdot (1 - S(t)) & \text { Reset-to-Zero } \\
H(t) \cdot (1 - S(t)) + (H(t) - V_{reset}) \cdot S(t) & \text { Linear Reset }
\end{array}\right.
\end{equation}

\begin{figure}[htb!]
	\center
	\includegraphics[width=0.8\linewidth]{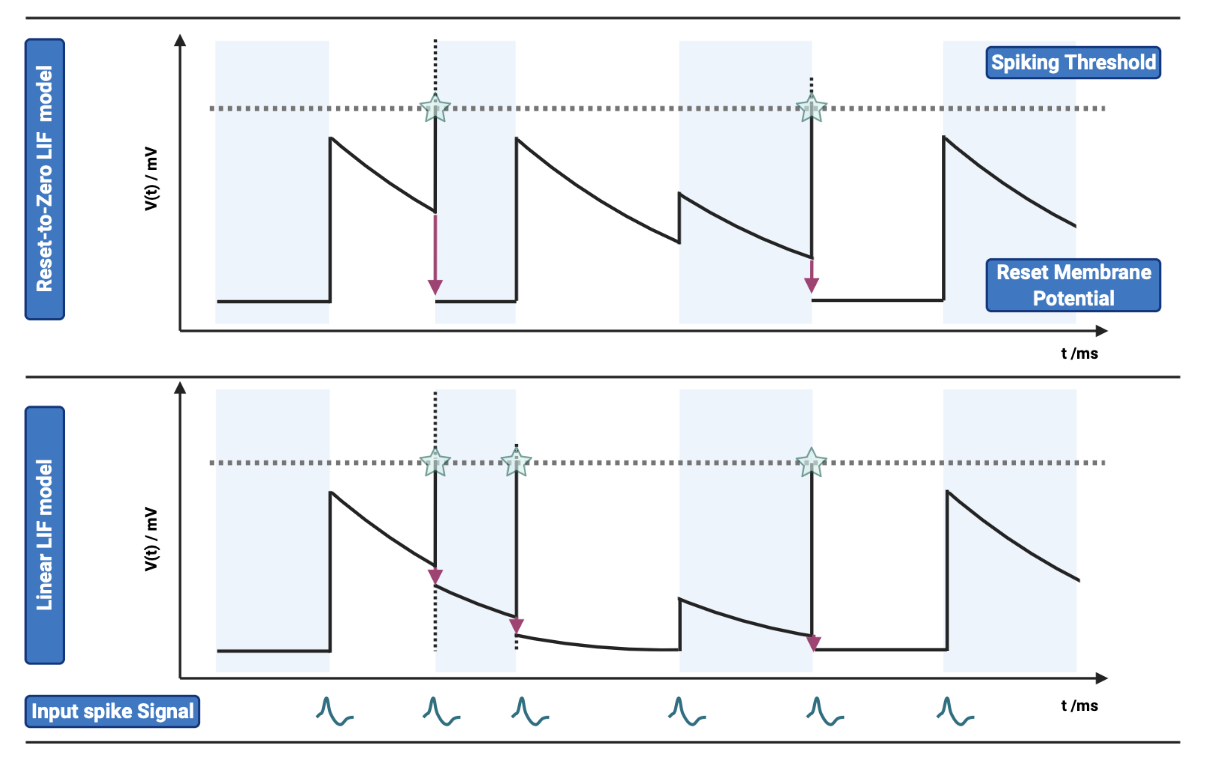}
	\caption{Comparison of membrane potential changes between LIF model and linear LIF model. \textbf{TOP}: 'Reset-to-Zero LIF model', the membrane potential will be reset to zero ($V_{reset} = 0 \ mV$). The subfigure shows the input action potential, membrane potential change, spiking threshold and output action potential from bottom to top. \textbf{Bottom}: 'Linear LIF model', membrane potential will subtracts the spiking threshold at the time when it exceeds the threshold.}
	\label{fig:continuouslif}
\end{figure}

 The LIF neuron model with 'Linear Reset' is named Linear LIF model. \cite{diehl2016conversion} and \cite{rueckauer2017conversion} analyzed the difference between these two MP reset modes and chose the linear LIF model for simulation. We analyze the two models from the perspective of physics and information theory and determine the advantages of the linear LIF model. For the first reset mode, the membrane potential of the normal LIF model does not satisfy the law of conservation of energy. There are two parts of membrane potential attenuations: 'leaky', the attenuations as the form of conductance in the circuit which keeps the nonlinear dynamic properties. The other part is that when the action potential is exceeded, the membrane potential exceeding the spike threshold will be lost directly, resulting in energy non-conservation. From the perspective of information, the linear LIF neuron model maintains the nonlinearity of the model and retains the completion of information to the greatest extent. The linear LIF model's membrane potential is shown in Fig.\ref{fig:continuouslif} comparison with the 'reset to zero' LIF model under the same input.

\subsection{Information transmission between LIF neuron models}
\begin{figure}[htb!]
	\centering
	\includegraphics[width=0.8\linewidth]{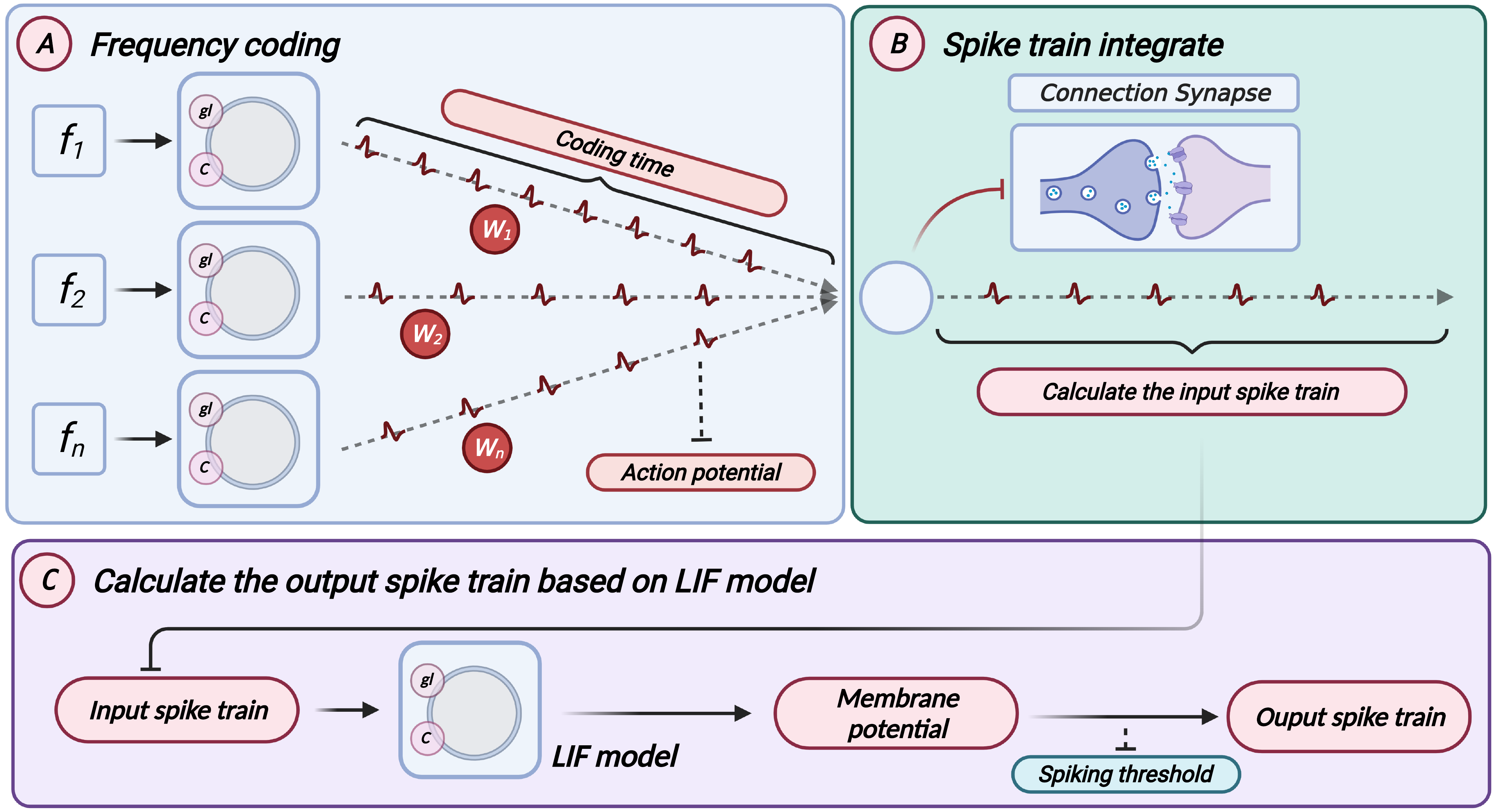}
	\caption{Information transmission of linear LIF model. A: All inputs are encoded as a spike train, multiplied by a weight. This is the information integration shown in the left part of the figure. B: Add all input spikes to get the input spike sequence of the neuron. C: The output action potential is obtained according to the input spike sequence through the LIF neuron's information processing mechanism.}
	\label{fig:calculate-the-output-spike-based-on-clif-model}
\end{figure}

The linear LIF neuron model has two steps in information transmission. The information is integrated by connection weight $\omega$, and then the data is processed and transmitted, as shown in Fig.\ref{fig:calculate-the-output-spike-based-on-clif-model}. We can see that the LIF model and linear LIF model are the same for subthreshold membrane potential changes in Fig.\ref{fig:continuouslif}. However, the resulting simulation proves that Linear LIF model is more similar to ReLU-AN model than LIF model.

\subsection{Mapping relationship between linear LIF and ReLU-AN}\label{mappingrelationship}

This subsection presents the mapping relationship between the linear LIF model and the ReLU-AN model from three aspects, i.e., weights, bias, and slope of the ReLU activation function. In comparison, the parameters of the linear LIF neuron are more biologically plausible, including the membrane capacitance $C_m$ and the membrane resistance $R_m$.
Here we give the parameter mapping we established in Tab.\ref{paramsrelationship}, which will be discussed in detail later in this paper. 

\begin{table*}[]
\centering
\caption{Parameter mapping between ReLU-AN model and linear LIF model}
\begin{tabular}{cccc}

\hline
\multicolumn{2}{c}{\textbf{Parameter of ReLU}}                                    & \multicolumn{2}{c}{\textbf{Params of linear LIF}}      \\[7pt] \hline
\multicolumn{1}{c|}{\textbf{Symbol}} &
  \multicolumn{1}{c|}{\textbf{Description}} &
  \multicolumn{1}{c|}{\textbf{Symbol}} &
  \textbf{Description} \\[7pt] \hline
\multicolumn{1}{c|}{$\omega$} & \multicolumn{1}{c|}{Connection weight}            & \multicolumn{1}{c|}{$\omega$}        & Synaptic weight \\[7pt] \hline
\multicolumn{1}{c|}{$b$} &
  \multicolumn{1}{c|}{Bias} &
  \multicolumn{1}{c|}{$\frac{ -\sum{\omega}}{R_m C_m \cdot \ln ( 1-\sum{\omega} /(V_{th} C_m))}$} &
   \\[7pt] \hline
\multicolumn{1}{c|}{$k$}      & \multicolumn{1}{c|}{Slope of activation function} & \multicolumn{1}{c|}{$1/{V_{th} C_m} $} &                 \\[7pt] \hline
\end{tabular}
\label{paramsrelationship}
\end{table*}

\subsubsection{Mapping of the weights}


We assume that the spiking frequency of the input signal is $f_j$ and the amplitude is 1, and then the signal can be expressed as:

\begin{equation}
	I_{l}=\sum_{i=1}^{n^l} \omega_{i}  \cdot \sum_{j=1}^{n}\delta(t-j\frac{1}{f_i})
\end{equation}
The $\omega_{i}$ is the synaptic weight between presynaptic neuron $i$ and post-synaptic neuron, $n^l$ represent the number of neuron in layer $l$, $j$ represents the $j_{th}$ action potential in the input spike train, $n$ is the number of action potentials and $T$ is the time windows of simulation.

Compared with the weight integration process in ANNs, we integrate the input signal $I_l$ in the time window $[0,T]$ and obtain:


\begin{equation}
 \int_{0}^{T} I_{l} dt = \sum_{i=1}^{n^l} \omega_{i}  \cdot \sum_{j=1}^{n} \int_{0}^{T}\delta(t-j\frac{1}{f_i})dt = T \cdot \sum_{i=1}^{n^l} \omega_{i} f_{i}
\end{equation}

This information integration mechanism is similar to the ReLU-AN model ($f = \sum\omega x$), except that for time-domain signals, the weight is reflected in the amplitude of the input action potential. We encode the weights in the amplitude of the spiking train but not in the Linear LIF model's conductance. Note that the encoding mechanism and information integration mechanism play a vital role in the network's information transmission process.

\subsubsection{Mapping of the bias}

The bias is an additional parameter in the ReLU model used to adjust the output along with the weighted sum of the inputs. Moreover, a learnable bias allows one to shift the activation function to either the right or the left. The neuron model's bias has a similar role with the threshold, determining whether the input activates the output. Based on Eq.\ref{ReLU} and the activation function, we know that the output of the neuron is equal to $0$ if $\sum_{i=1}^{n} w_{i} x_{i}<-b$.

Similarly, it cannot be fired if the input frequency is less than a threshold. According to the spike excitation rules of Linear LIF neurons, the action potential is generated when the membrane potential reaches the spiking threshold. If the integrated spike train can excite an action potential within the time window $t_{\omega}$ (set to 4 s in the simulation here), it should satisfy

\begin{equation}
	V(T=n / f_{i n})>V_{th}
\end{equation}
where $T$ refers to the time window, and $n$ is the number of action potentials within the time window at the current frequency. According to Eq. \ref{CLIF},  we can get the relation between the number of action potentials and the other parameters.

\begin{equation} \label{mappinggl}
	\sum_{i=1}^{n^l} \omega_i f_{i}>\frac{-\sum_{i=1}^{n^l} \omega_{i}}{\tau_m \cdot \ln (1-\sum_{i=1}^{n^l} \omega_{i}/(V_{th} C_m) \cdot(1-e^{-T/\tau_m}))}
\end{equation}

The Linear LIF model can excite the action potential in the time window only if the inequality is satisfied, that is when the input frequency is greater than a value determined by the parameters of the Linear LIF model. This mechanism has the same effect as the bias, which filters the input signal before the information processing. Bias of the ReLU-AN model also can be expressed as a function of membrane resistance $R_m$ and membrane capacitance $C_m$. The existence of membrane resistance $R_m$ endows the Linear LIF model with nonlinearity, which is very important for neural networks.

\subsubsection{Mapping of activation function}

The activation function determines the relationship between input and output after integration. We focus on the non-negativity and linear relationship of ReLU. For non-negativity, the Linear LIF model's output is based on the number of action potentials, a non-negative value. So, for the input-output relationship where the input is greater than 0. The relationship between integrated input and output spike frequencies of the LIF neurons can be expressed as 

\begin{equation}
	f_{o}=\frac{f_{ {in }}}{\lfloor n\rfloor} \quad\{n \mid V(n /f_{in}) \geq \theta\}
\end{equation}
where $f_{in}$ is the frequency of input signal, $f_o$ is the frequency of the output signal, and $n$ is the minimum number of input spikes capable of firing an action potential within a time window, defined by Eq.\ref{CLIF}.

We rounded up the number of input spikes and approximated the relation between the input and output frequency as

\begin{equation}
	f_{o}=\tau_m[\frac{[(V_{th} C_m)/\sum_{i=1}^{n^l} \omega_{i}](1-e^{-1/{f_{i} \tau_m}})}{1-[(V_{th} C_m)/\sum_{i=1}^{n^l} \omega_{i}](1-e^{-1/{f_{i} \tau_m}})}]
\end{equation}

Based on the action potential frequency in biological neurons, we assume that the input frequency $f_{in}$ satisfies $-1 /(f_i \tau_m) \approx 0$. According to the approximation formulas $\ln (1+x) \approx x$	 and $e^x \approx 1+x$, the above formula can be simplified to

\begin{equation}\label{mappingcm}
	f_{o}=\frac{1}{V_{th}C_m} \cdot \sum_{i} \omega_{i=1}^{n^l} f_{i}
\end{equation}

The input frequency $f_{i}$ is proportional to the output frequency $f_o$, as shown above. If $1/(V_{th}C_m)= 1$, we can conclude that this function is equivalent to the ReLU, in the case of the same input frequency. No matter how the parameters of the LIF model change, this proportional relation remains true. Given the spiking threshold, the LIF model introduces nonlinearity into the network. Simultaneously, because of frequency coding, the input frequency cannot be less than 0, which increases the sparsity of the network. We conclude that the LIF model can be equivalent to the ReLU under certain parameter mapping principles.

\subsection{Definition of model equivalence}

In this subsection, we define the equivalence in two aspects, i.e., structural equivalence and behavioral equivalence (shown in Fig.\ref{fig:definition-of-model-equivalence}):

\begin{figure}[htb!]
	\centering
	\includegraphics[width=0.8\linewidth]{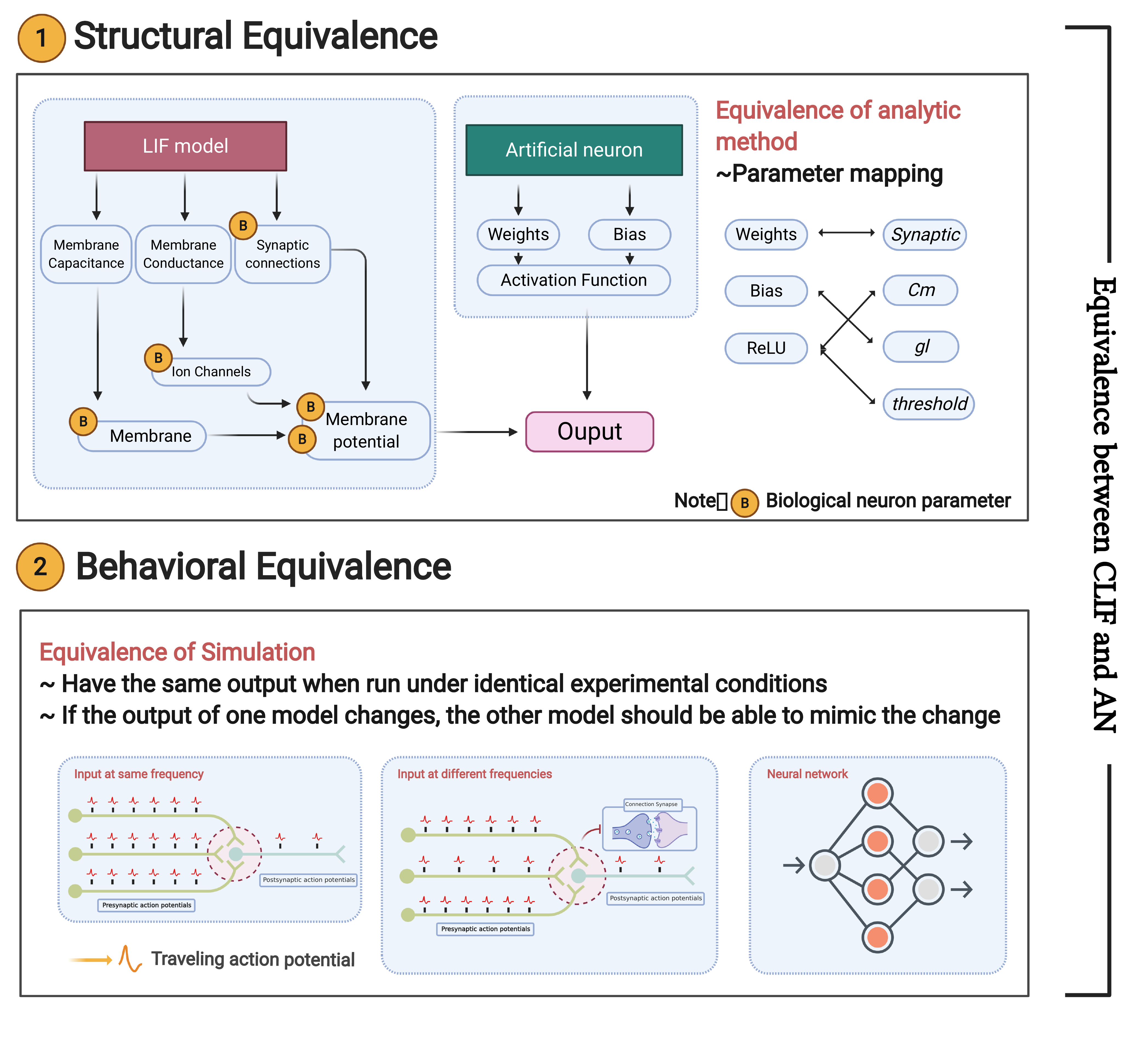}
	\caption[]{\textbf{Equivalence between Linear LIF/SNN model and ReLU/ANN}: structural equivalence and behavioral equivalence.}
	\label{fig:definition-of-model-equivalence}
\end{figure}

\begin{itemize}
	\item \textbf{Structural equivalence} is mainly reflected in the structures of the ReLU-AN model and LIF model. The parameters of the two models should have a mapping relationship represented by a transformation function $R$. $R$ can be described as a binary relation satisfying reflexive ($\mathrm{xRx}$), symmetrical $\mathrm{xRy} \Rightarrow \mathrm{yRx}$), and transitive properties ($(\mathrm{xRy} \wedge \mathrm{yRz}) \Rightarrow \mathrm{xRz}$). We will present a perfect parameter mapping between Linear LIF/SNN (model A) and ReLU/ANN (model B) in section \ref{mappingrelationship}. 
	\item \textbf{Behavioral equivalence} focuses on the functional equivalence of the two models, requiring that model A can complete the functions of model B, and vice versa. We define behavioral equivalence as: “Model A and Model B have the same output if run under identical experimental conditions. Given a parameter mapping rule, there always exists a small error bound $\varepsilon$ that $\left\|\mathrm{F}_{A}(x)-\mathrm{F}_{B}(x)\right\| \leq \varepsilon$ can be guaranteed for any valid input $x$, where $F_A$,$F_B$ denotes the function of model A and model B.”
\end{itemize}

\section{Experiments and Analysis}

\begin{figure*}[h!]
	\centering
	\includegraphics[width=0.8\linewidth]{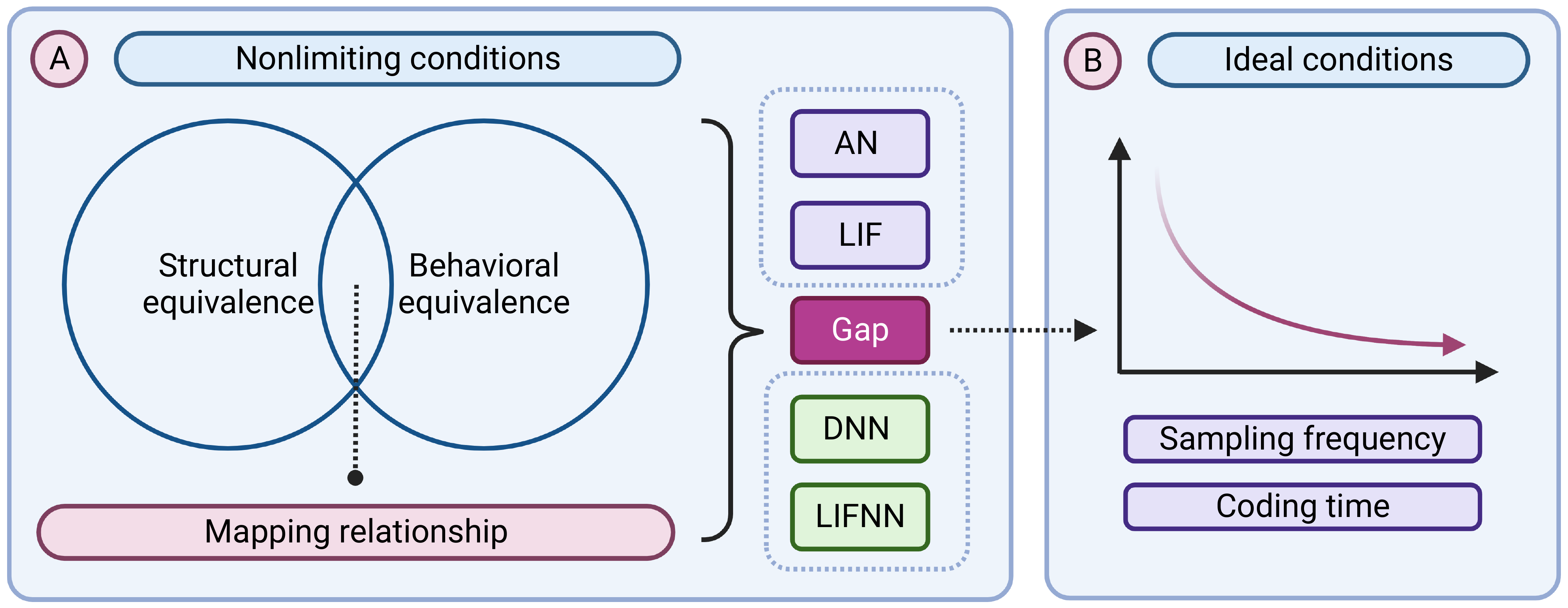}
	\caption{The framework of the proof of equivalence between Linear LIF model and ReLU model}
	\label{fig:the-equivalence-of-clif}
\end{figure*}

In this section, we demonstrate the equivalence of LIF/SNN and ReLU-AN/DNN model and the advantages of the linear LIF model compared to the reset-to-zero LIF model. As shown in Fig.\ref{fig:the-equivalence-of-clif}, it mainly includes: 1. Verify the structural and functional equivalence of LIF/SNN and ReLU/DNN through simulation. 2. Reduce the simulation error by increasing the sampling frequency and coding time, demonstrating a convergence towards ideal conditions.  

The simulation experiment in this section is mainly divided into two parts: 

\begin{enumerate}
	\item Simulation 1: Proof of structural equivalence
	\begin{enumerate}
		\item Compare the linear LIF model with the ReLU-AN model (with bias) when the input signal frequencies are the same.
	\end{enumerate}
	\item Simulation 2: Prove of behavioral equivalence
	\begin{enumerate}
		\item Compare the linear LIF model and the ReLU-AN model (with bias) under the condition that the two input signal frequencies are different.
		\item Compare the linear LIF model and the ReLU-AN model (with bias) under the condition that the input signal frequencies are different.
		\item Compare LIFNN and DNN (without bias) based on face/motor data set and MNIST and CIFAR10 data set.
	\end{enumerate}
\end{enumerate}



\subsection{Information coding}

Neuronal coding is a key step of the simulation. Neurons use sequences of action potentials, which can be considered as a point process, to carry information from one node to another in the brain. This spiking train can be considered an element of neural coding. The shape and duration of the individual spikes generated by a given neuron are very similar, so we think that the spike train can be described as a train of one-or-none point events in time \cite{kostal2007neuronal}.

The information received by the LIF neuron is a series of spike sequences, so the original data needs to be encoded into a spike train \cite{RICHMOND2009137}. Many coding methods are used in biological neuron models, such as one-dimensional coding and sparse coding mechanism. We assume that the neuron encodes the information as a spike frequency and encodes it according to the following rules:

\begin{equation}
	I_{l}^{i}= \omega_{i}  \cdot \sum_{j=1}^{n}\delta(t-\frac{j}{x_{i}}), \quad \frac{n}{x_{i}} <= T_w
\end{equation}
where $x_i$ ($x_i \in [0,1]$) is the input. We encode the input as a series of spike sequences with a fixed frequency, which is proportional to the input data. The encoding time (time window) is $T_w$, sampling frequency is the maximum frequency of encoding $R_{max}=1/ \Delta t$. In our experiment, we set the minimum time unit $\Delta t$ as 0.01s and the time window $T_w$ as 3s. That is, a pixel is encoded into a sequence with a length of 300 (time steps).

\subsection{Simulation1: Proof of structural equivalence}\label{section_simulation1}

\begin{figure}[htb!]
	\center
	\includegraphics[width=0.8\linewidth]{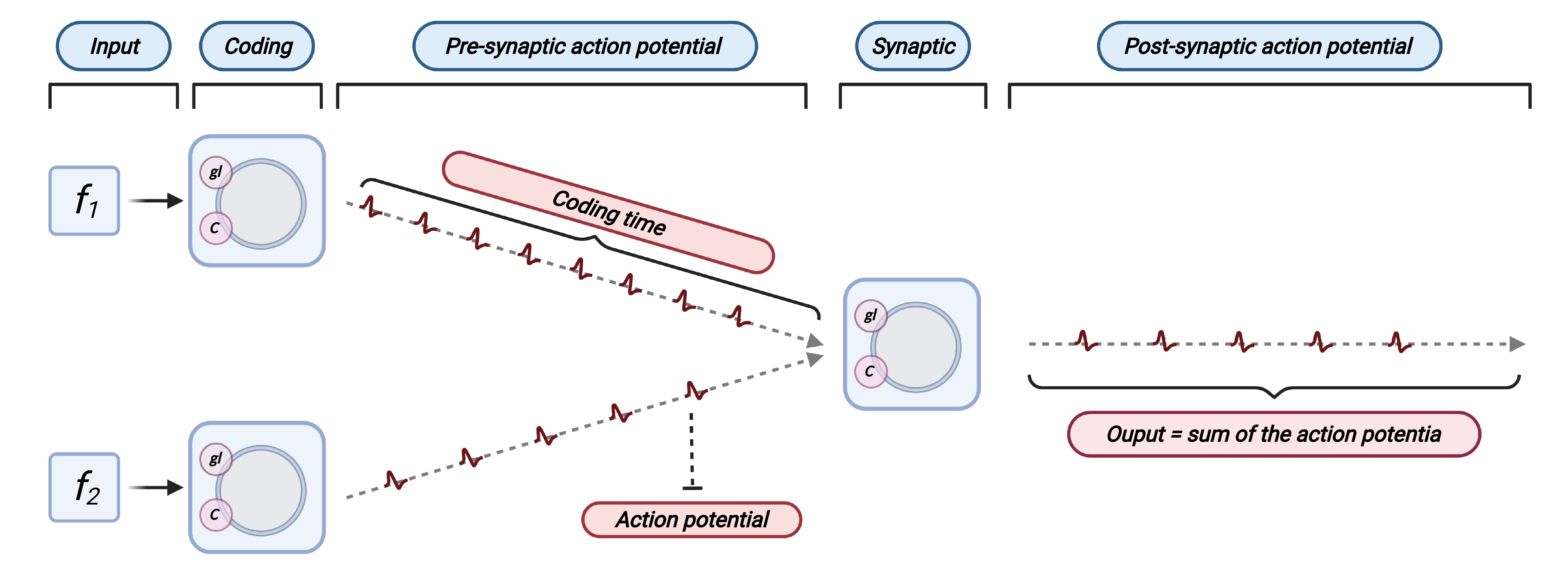}
	\caption{The structure of the simulation. The whole structure consists of three neurons, two of which are used for information coding and serve as presynaptic neurons; the third LIF neuron is used to process the input of the first two neurons and produce an output spike train. There are two synaptic structures in this structure.}
	\label{fig:the-structure-of-simulation}
\end{figure}

In this subsection, we examine the dynamics of the linear LIF model for verifying the proposed parameter mapping. The simple structure consists of a single LIF neuron and two synaptic, shown in Fig.\ref{fig:the-structure-of-simulation}.

Firstly, we consider the simplest case, and the LIF model has two input spike trains from the presynaptic neurons with the same frequency. When calculating the output of the linear LIF model and the reset-to-zero LIF model, we transfer the weights and obtain the values of membrane capacitance and membrane conductance according to the mapping relationship, so that the input and output relationship of the Linear LIF model is the same as the ReLU model. The parameters setting is shown in Tab.\ref{tab:my-table3}. 

\begin{table}[htb!]
	\centering
	\caption{Parameters setting of simulation 1}
	\label{tab:my-table3}
	\begin{tabular}{l|l|l|l|l|l|l|l}
		\hline
		Params & $f_{in}$      & $\omega$    & $b$   & $gl$ & $C_m$ & $V_{th}$ \\ \hline
		Value  & $[1,30]$ Hz & $[0.3,0.2]$ & -2.16 & 3.0  & 1.0 & 1.0      \\ \hline
	\end{tabular}
\end{table}

\begin{figure*}[htb!]
	\centering
	\includegraphics[width=1\linewidth]{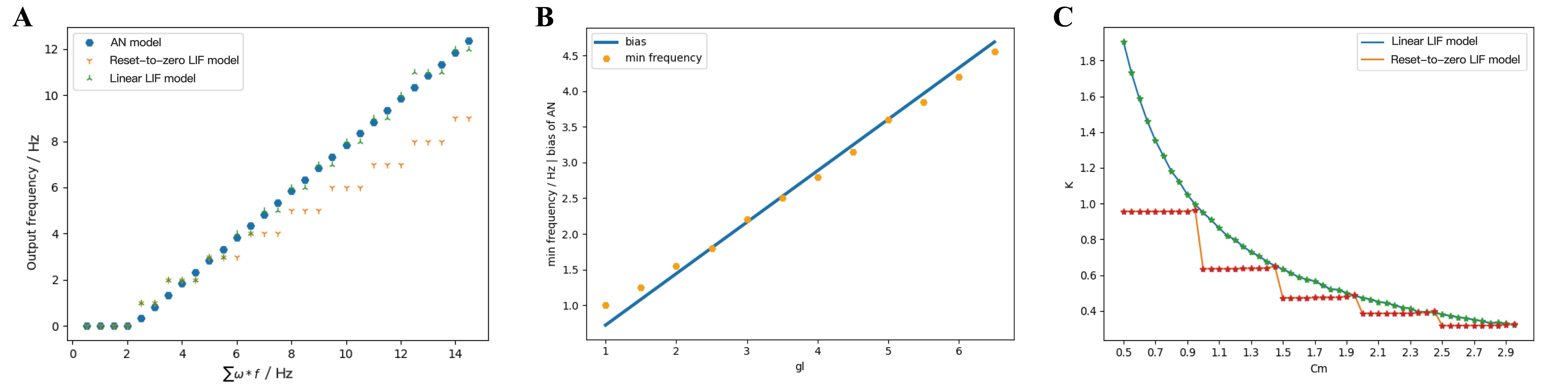}
	\caption{\textbf{A.} The output of ReLU, LIF and Linear LIF when the input signals have same frequency. The abscissa is the input frequency and the ordinate is the output frequency. The blue point represents the output of ReLU, the orange represents the output of the reset-to-zero LIF model, and the green point represents the data of the linear LIF model. \textbf{B.} The relationship between min frequency and membrane conductance. \textbf{C.} The relationship between slope of activation function and membrane capacitance. }
	\label{biasandcm}
\end{figure*}

We recorded the output frequency/pixel of ReLU-AN model, Linear LIF model, and Reset-to-zero LIF model under the same input, shown in Fig.\ref{biasandcm}. Fig.\ref{biasandcm}A shows that the output of linear LIF model is almost equal to the output of ReLU-AN model, but reset-to-zero LIF model has a gap when the input frequency is large. The slope of the relationship between input and output, which is calculated by $1/(C_m V_{th} )$, is equal to that of ReLU-AN model. Besides, the minimum input frequency that can excite an action potential is consistent with the bias of ReLU-AN model.

Fig.\ref{biasandcm} B shows the relationship between the membrane conductance $gl$ and the bias. Apparently, the $gl$ is proportional to the minimum input frequency which can excite the action potential and good agreement with the Eq.\ref{mappinggl}. By changing membrane capacitance $C_m$ and fixing other simulation parameters, we can get the relationship between the slope of activation function and membrane capacitance $C_m$. Fig.\ref{biasandcm}C illustrates the relationship of linear LIF (green) with respect to the reset-to-zero LIF model (red). The linear LIF model is more consistent with Eq.\ref{mappingcm}, while the LIF model is a step-like descent. 

Based on this simulation, we verified the mapping relationship between $C_m$ and the slope of input-output line, $gl$, and the bias. So that the structural equivalence is proved that the parameters of linear LIF  model and ReLU-AN model can be one-to-one corresponded.

\subsection{Simulation2: Proof of behavioral equivalence}\label{biassimulation}

In this subsection, we prove that the behavioral equivalence is valid under various conditions. If the difference between the linear LIF model and the ReLU-AN model is within the error range, we believe that the behavioral equivalence is valid. In this simulation, we get the structural equivalence, that is, the parameter mapping relation is applicable in all the above cases.

\subsubsection{Experiments for two input spiking trains}

We use the same structure and coding algorithm as section \ref{section_simulation1}. We map the parameters of ReLU-AN model to the linear LIF model according to the parameter mapping relationship. When the input spiking trains have different frequencies, the output signal of linear LIF model is not a periodic spiking train. So we count the number of action potentials in the output spiking train and divide it by the time windows to get the output frequency. The params of the simulation are given in Tab.\ref{tab:my-table2}.

\begin{table}[htb!]
	\centering
	\caption{Parameters setting of simulation 2}
	\label{tab:my-table2}
	\begin{tabular}{l|l|l|l|l|l|l|l}
		\hline
		Params & $f_{in}$      & $\omega$    & $b$   & $gl$ & $C_m$ & $V_{th}$  \\ \hline
		Value  & $[1,60] \ Hz$ & $[0.3,0.2]$ & -2.16 & 3.0  & 1.0 & 1.0       \\ \hline
	\end{tabular}
\end{table}

\begin{figure*}[htb!]
	\centering
	\includegraphics[width=1\linewidth]{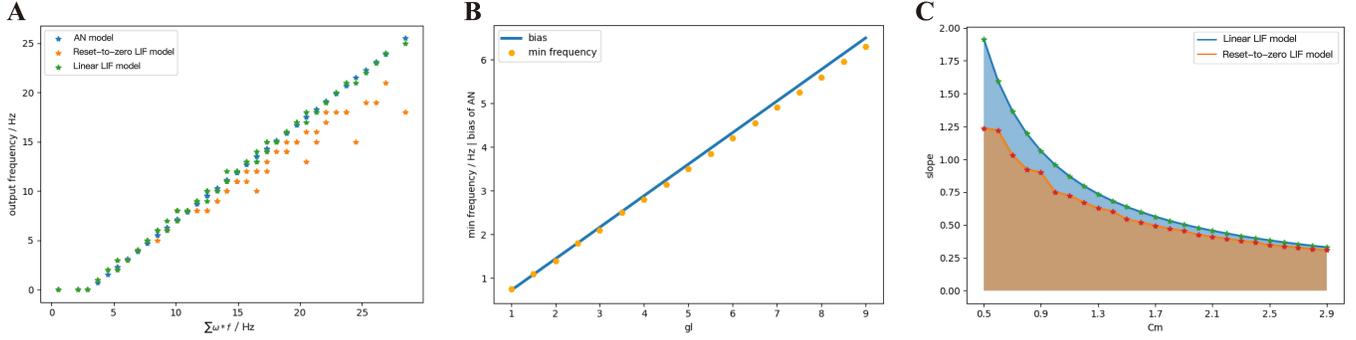}
	\caption{\textbf{A} The output of ReLU-AN model, reset-to-zero LIF model and linear LIF model when the input signals have different frequencies. \textbf{B} The relationship between membrane conductance $gl$ and the bias. \textbf{C} The relationship between membrane capacitance $C_m$ and slope of activation function.}
	\label{simulation2}
\end{figure*}

The output of ReLU-AN model, linear LIF model, and reset-to-zero LIF under the same condition, as shown in Fig.\ref{simulation2}A. The linear LIF model's output can fit well with the output of ReLU-AN model, while there is an error between it and reset-to-zero LIF model. We perform additional experiments to explore the relationship between $gl$ and bias. Fig.\ref{simulation2}B illustrates the bias change of the ReLU-AN model (blue) with respect to the minimum frequency of the linear LIF model. We observe that the minimum frequency fits well with the bias, which indicates the correctness of behavioral equivalence. Fig.\ref{simulation2}C shows the relationship between $C_m$ and slope of activation function, where the blue line represents the linear LIF model and the orange line represents the reset-to-zero LIF model. According to the mapping relationship, the slope of the input-output curve should be inversely proportional to the membrane capacitance $C_m$. We can conclude that the linear LIF model is more consistent with the Eq.\ref{mappingcm}, and the slope of the input-output function can be adjusted to 1 through parameter tuning.

\begin{figure}[htb!]
	\centering
	\includegraphics[width=0.6\linewidth]{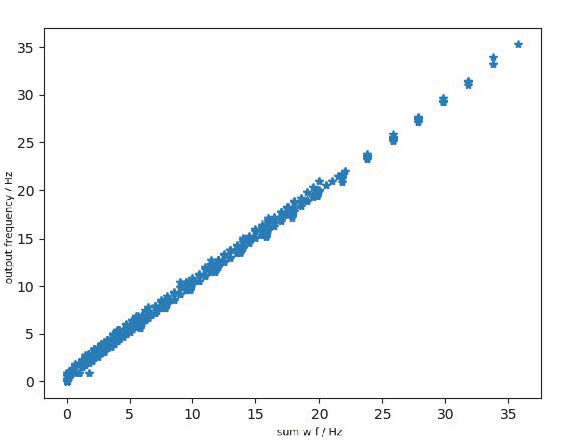}
	\caption{The output of ReLU-AN model, reset-to-zero LIF model and linear LIF model when the input signals have three different frequencies.}
	\label{fig:simulation2inputandoutputmuti}
\end{figure}

Lastly, we discuss the behavioral equivalence of the linear LIF model and ReLU-AN model under the condition that the model has three input signals with different frequencies. As shown in Fig.\ref{fig:simulation2inputandoutputmuti}, the output of linear LIF is the same as the output of ReLU-AN model within a certain margin of error. In this case, we proved that the behavioral equivalence is valid under this condition.

\subsection{Experiments for full connected architectures}

In this section, we will analyze a case of a three-layer neural network. one is a pre-trained ANN based on training dataset, and another is the weights converted linear LIF neural network (LLIFNN) based on the pre-trained ANN. We will analyze the middle-layer output, the classification accuracy of the test dataset, and the influence of parameters on the neural network.

\begin{figure}
	\centering
	\includegraphics[width=0.9\linewidth]{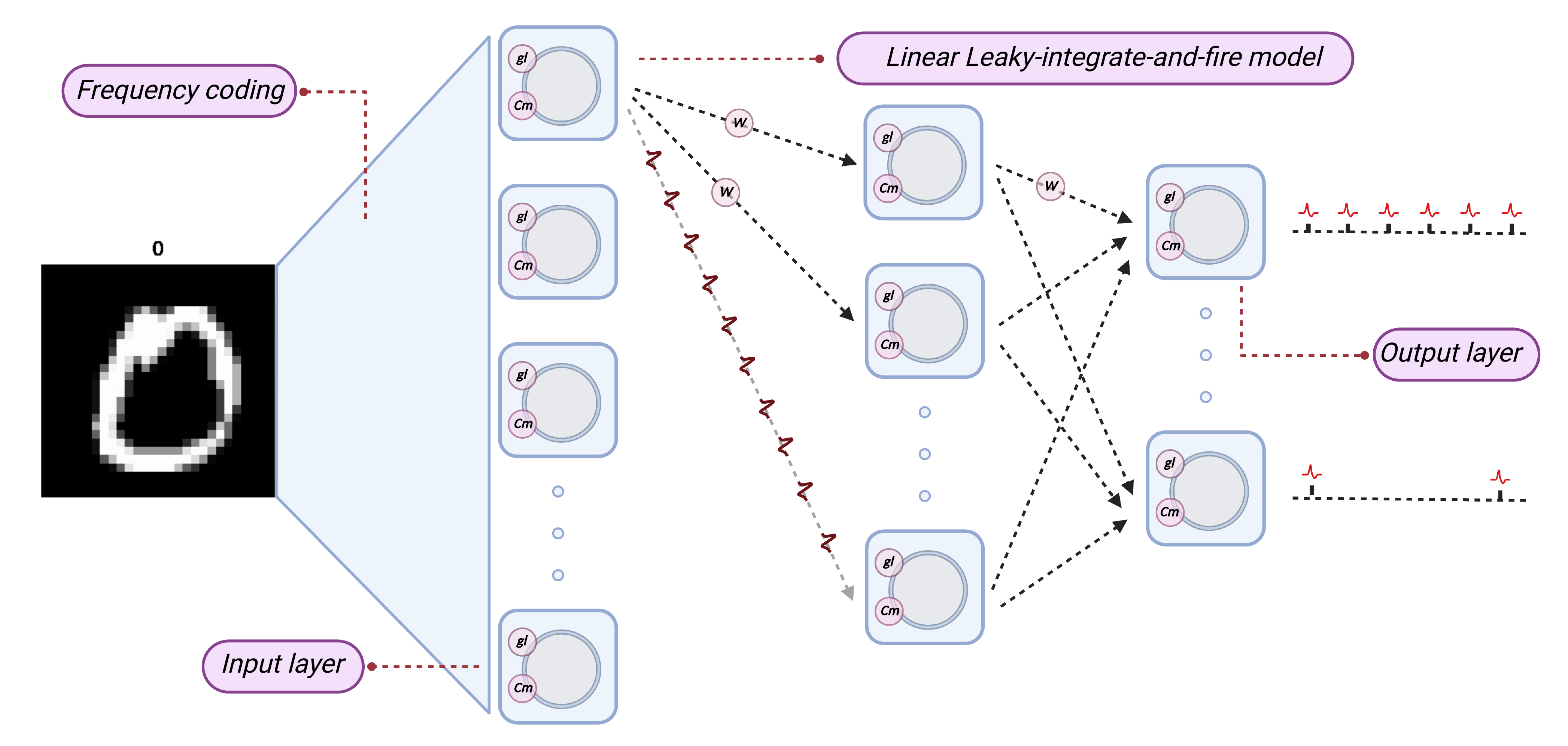}
	\caption{\textbf{The structure of LIFSNN}. The network structure of SNN is the same as that of ANN, and the nodes are replaced by Linear LIF model. The subscript of the node with the largest number of action potentials in the output layer is taken as the output of the network.}
	\label{fig:network-of-simulation-3}
\end{figure}

We established an equivalent LLIFNN according to the proposed parameter mapping. We set the parameters of the Linear LIF neurons to fixed values, such as the membrane capacitance $C_m$ and spiking threshold $V_{th}$, and mapping the weights of trained networks to LLIFNN. Since the bias of nodes in ANN is set to zero, so we also set the membrane capacitance $C_m$ to a fixed value. We use the frequency coding and mark the subscripts of the node with the most action potentials in the output layer as the label. The structure and information processing algorithm are shown in Fig.\ref{fig:network-of-simulation-3}.

The structural equivalence and behavioral equivalence constitute the equivalence between networks. We use correlation coefficient \cite{meng1992comparing} to quantify the functional equivalence of the network, which is also the most common indicator of similarity. The correlation coefficient is introduced as a metric of the similarity of neural information. It is used in statistics to measure how strong a relationship is between two random variables. The correlation coefficient between $x$ and $y$ is

\begin{equation}
	\rho_{x, y}=\frac{n(\sum x y)-(\Sigma x)(\Sigma y)}{\sqrt{[n \Sigma x^{2}-(\sum x)^{2}][n \Sigma y^{2}-(\Sigma y)^{2}]}}
\end{equation}

\subsubsection{Face/Motorbike dataset}\label{facedataset}

We firstly evaluate LLIFNN based on the face and motorbike categories of the Caltech 101 dataset, including 400 training pictures and 464 test pictures. We unified the image size to 130*80 pixels. We trained and tested a three-layer fully connected neural network without bias by presenting 20 epoch training set. The data of size 130*80 is stretched into a 10400-demensional vector and input to the neural network. The 600 neurons in the hidden layer integrate the input data of the input layer and pass them to the classification layer (of 2 neurons) through the ReLU function. The structure of LLIFNN is the same as ANN. Take the output of ANN as the label of test data, classification of LLIFNN based on face-moto dataset can achieve 99.75\%. The confusion matrix is shown in Tab.\ref{tab:face-moto-dataset}.

\begin{table}[htb!]
	\centering
	\caption{Confusion matrix of face/motorbike dataset}
	\label{tab:face-moto-dataset}
	\begin{tabular}{|c|c|l|l|}
		\hline
		\multicolumn{2}{|c|}{\multirow{2}{*}{Confusion matrix}} & \multicolumn{2}{c|}{Predict}                          \\ \cline{3-4} 
		\multicolumn{2}{|c|}{}                                  & \multicolumn{1}{c|}{Face} & \multicolumn{1}{c|}{Moto} \\ \hline
		\multirow{2}{*}{Real}               & Face              & 200                       & 0                         \\ \cline{2-4} 
		& Moto              & 1                         & 199                       \\ \hline
	\end{tabular}
\end{table}

\subsubsection{MNIST dataset}\label{mnistsubsection}

The MNIST dataset is a large database of handwritten digits that contains ten object categories. It has 60000 training images and 10000 test images. The MNIST dataset is a good benchmark to show and prove the behavioral similarity between LLIFNN and ANN. We use the network with the same structure as the network in section \ref{facedataset}. We will analyze the behavioral similarity from two aspects: the output of the middle layer and output layer, respectively. 

\begin{figure}[htb!]
	\centering
	\includegraphics[width=1\linewidth]{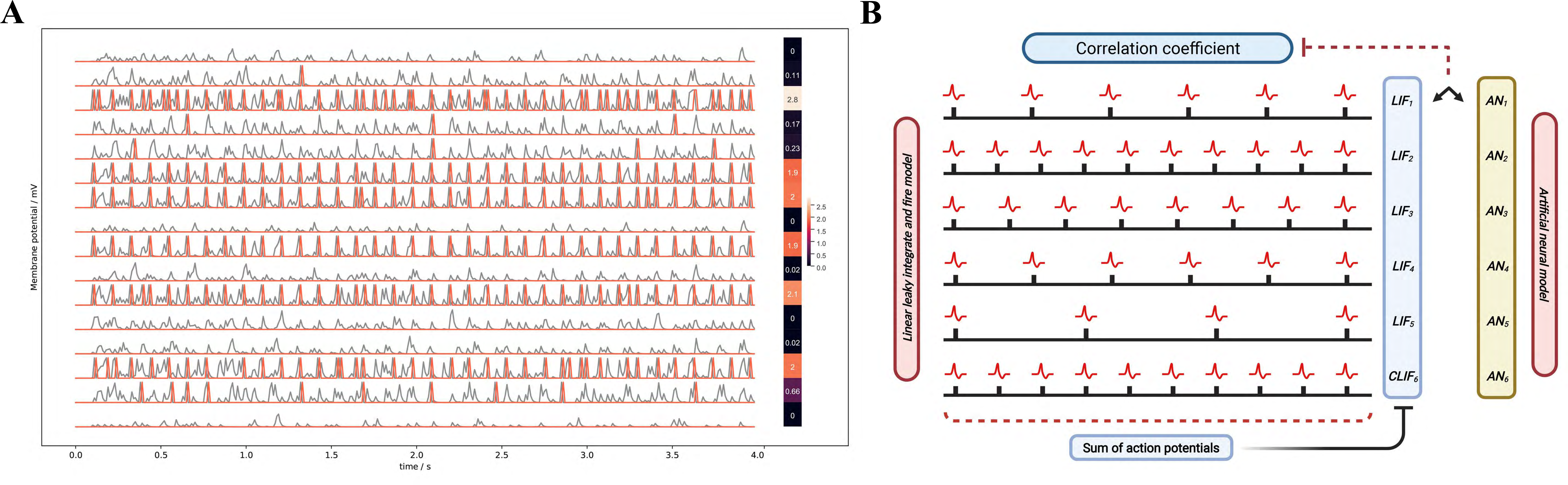}
	\caption{\textbf{A.} The membrane potential of LIF neurons. The left curve shows the state of 16 LIF neurons. The first heat map on the right shows the number of spikes of LIF neurons as arranged in a map. Each data corresponds to the LIF membrane potential data on the left. The other heat map is the same but for the ReLU neurons. \textbf{B.} The framework of calculating the correlation coefficient.}
	\label{fig:11}s
\end{figure}

Here we show the output spike trains of the middle layer and mark the change of membrane potential and the action potentials. In Fig.\ref{fig:11} A, the grey lines represent the change of membrane potential and the red lines represent the action potential. When the membrane potential reaches the spiking threshold, the post-synaptic neuron will excite an action potential.

Firstly, we compare the output of the middle layer based on the correlation coefficient analysis. The framework of calculating the correlation coefficient is shown in Fig.\ref{fig:11}B. We convert the 60 series (the number of nodes in the middle layer) of spiking trains in the middle layer of LLIFNN into a 60*1 vector, named vector A. The output calculation method of Linear LIF is the same as the method mentioned in the previous chapter. With the middle layer of ANN, a 60*1 vector named vector B. We can calculate the correlation coefficient of the two vectors to quantitatively analyze the correlation between the two outputs.

\begin{figure*}[htb!]
	\centering
	\includegraphics[width=1\linewidth]{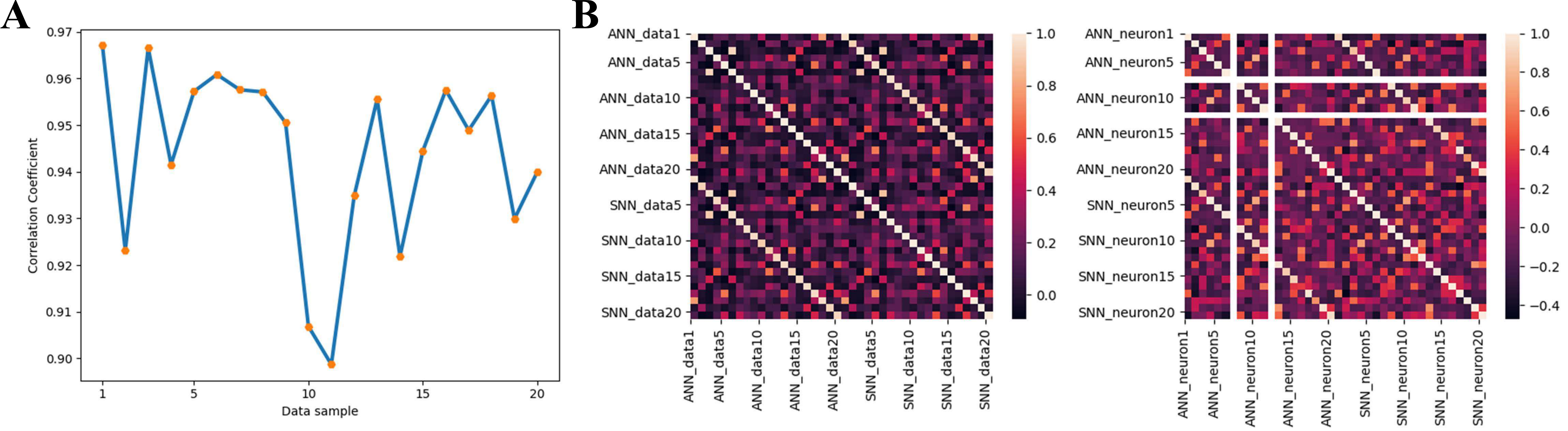}
	\caption{\textbf{A.}Correlation of the output of the two models for different data. \textbf{B.} The correlation coefficient matrix along data dimension (left) and neuron dimension (right).}
	\label{Correlation_coefficient}
\end{figure*}

Fig.\ref{Correlation_coefficient}A shows the correlation coefficients for 20 groups of data, where the abscissa is the data table and the ordinate is the correlation coefficient. If the correlation coefficient is set to greater than 0.8, it can be determined that the two variables are strongly related. The orange line is the change in the correlation coefficient of the output signals of two models based on 50 groups of data, and the blue dotted line is the average value of the correlation coefficient. We mark in the image shows that the maximum correlation coefficient is 0.97, the minimum is 0.90, and the average is 0.94. It quantitatively validates the similarity between the outputs of middle layer.

We assume that the behavioral equivalence of neural networks can be reflected in two aspects: 1. For the same data, the outputs of ANN and SNN interneurons are equivalent. 2. The output of the neuron at the same location is equivalent under different data input situations. Fig.\ref{Correlation_coefficient}B shows the correlation coefficient matrix in data dimension and neuron dimension, corresponding to the two aspects respectively. The two correlation coefficient matrices are Symmetric Matrices and can be divided into four quadrants with the center point as the origin of the Darwin coordinate system.

We denote by $x$ the output of the DNN, by $y$ the output of the SNN, and by $\rho (x,y)$ the correlation coefficient. Based on 20 groups of test data, the correlation coefficient matrix between the output of the Linear LIF model and the output of the ReLU model in the middle layer is shown on the left side of Fig.\ref{Correlation_coefficient}B. The first quadrant is the correlation coefficient matrix $\rho (x,x)$ of ReLU vs. ReLU among different data, the second quadrant shows the correlation coefficient matrix $\rho (x,y)$ between the output of the LIF model $x_{[600,20]}$ and the output of the ReLU model $y_{[600,20]}$, and the third quadrant is the correlation coefficient matrix $\rho (y,y)$ of LIF vs. LIF among different data. We call this the data-dimension correlation coefficient matrix. In other words, for a single data, the output of LIF is more similar to the output of ReLU under the same input than the output of the ReLU model under different inputs. Next, we analyze the neuron-dimension correlation coefficient. We analyze the correlation coefficient matrix $\rho (x_{i,j},y_{i,j} )$, where $i$ is the subscript of the data and $j$ is the subscript of the neuron, shown in the right part of Fig.\ref{Correlation_coefficient}B. Because there are two LIF neuron that did not fire an action potential in all the test data, its variance in 20 sets of data is $0$, which will cause an error in the calculation of the correlation coefficient matrix. So we ignore this neuron and only consider 19 neurons. The first quadrant is the correlation coefficient matrix of ReLU vs. ReLU, the second quadrant shows the correlation coefficient matrix of Linear LIF vs. ReLU, and the fourth quadrant is the correlation coefficient matrix of Linear LIF vs. Linear LIF, but all for different neurons. We can see that the features represented by an Linear LIF neuron are equivalent to the features extracted by the corresponding ReLU neuron.

\begin{figure*}[hbt!]
  \includegraphics[width=1\linewidth]{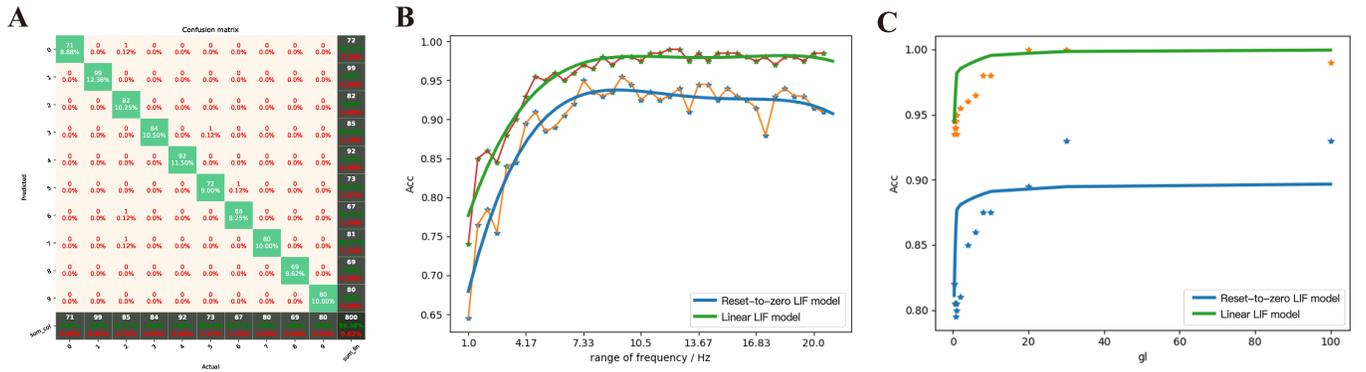}
  \caption{\textbf{(a)} The confusion matrix between the output of the network built by LIF/SNN and the predicted value of ReLU/DNN. \textbf{(b)} Comparison of classification accuracy of MNIST between LIFNN and LIFNN by changing the range of coding frequency. \textbf{(c)} Comparison of classification accuracy of MNIST between LIFNN and LIFNN by changing membrane conductance.}
  \label{Correlation_coefficient1}
\end{figure*}

Through the calculation of the correlation coefficient matrix, we have illustrated the behavioral equivalence of the hidden layer. To further illustrate the behavioral equivalence of network, we calculated the classification accuracy of LIF/SNN, using the output label of ANN as the benchmark. Based on the classification accuracy of MNSIT dataset, not only the equivalence of the output layer can be verified, but also the equivalence of the entire network can be proved. We used the subscript of node with the largest number of action potential as the final classification label. A confusion matrix of the SNN network classification results against that of the DNN classification results is depicted in Fig.\ref{Correlation_coefficient1}A. The overall accuracy reaches 99.38\%, indicating that under the proposed parameter mapping, the LIF/SNN can achieve similar classification results as its equivalent ReLU/DNN.

The LIF model is widely used in other articles focus on SNN, and the Linear LIF model is modified based on the LIF model. With an understanding of the difference between the LIF model and the Linear LIF model, what advantages does the Linear LIF model have in the network? We change the following two parameters and analyze the gap between the classification accuracy of the two networks MNIST to quantitatively compare the two models:

\begin{itemize}
	\item The range of coding frequency. We know that the data range of MNIST is $[0,1]$. Based on the frequency coding algorithm, we can encode the real number into a spike train within the coding time. Since the weight which is transformed from ANN is small (the maximum value is around 0.1), we give a parameter to map the coding range to $[0,k]$.
	\item Membrane conductance. In the parameter mapping relationship, $gl$ maps to the bias of ReLU, but in this ANN we set the bias to $0$. The parameter $gl$ determines the membrane potential attenuation of the LIF and Linear LIF models in the time domain, so we also make it as a variable to compare the LIF model and the Linear LIF model. 
\end{itemize}

Fig.\ref{Correlation_coefficient1}(b) shows the trend of network classification accuracy based on the two models with the range of coding frequency. We can see that as the range of coding frequency increases, the recognition accuracy of the two networks is increasing. However, under the same parameters and weights, the recognition accuracy of the LIFNN is 7\% higher than that of LIFNN. At the same time, as $gl$ increases, the accuracy of the two networks also increases, shown in Fig.\ref{Correlation_coefficient1}(c). It can be concluded that the Linear LIF model has advantages in network construction, compared with the LIF model. However, there are still some remaining problems. For example, according to the parameter mapping relationship, the smaller the membrane conductivity parameter, the higher the equivalence between the Linear LIF model and the ReLU model. For the network structure built by multiple Linear LIF models, the larger the film capacitance parameter, the higher the classification accuracy compared to ANN. We believe that the membrane conductance has a complicated relationship with the frequency coding range. We will explore in the follow-up work and believe that the convolutional SNN will eliminate this problem after adding bias.

Through the comparison of middle layers of LIFSNN and ANN, the comparison of classification accuracy of MNIST data set, and the comparison of the classification effect of Linear LIFSNN and LIFSNN, we proved that LIFNN and ANN are behavioral equivalent at the network level, which confirms the behavioral equivalence of Linear LIF model and the ReLU model.

\subsection{Experiments for convolutional architectures}

The previous section proved the behavioral equivalence based on fully connected neural networks (FCNNs). However, FCNNs with a large number of parameters require a longer training time and overfit the training dataset. Compared with FCNNs, convolutional neural networks (CNNs) are quite effective for image classification problems and have been applied for various learning problems. The convolutional layer, the main component of CNNs, is different from full-connection layers. To bridge the gap between Deep learning and SNNs, we verified that SNNs with the convolutional structure could also complete the task of CNN, based on the equivalence between the Linear LIF model and ReLU-AN model.

\subsubsection{Spiking convolutional layer}

\begin{figure}[h!]
	\centering
	\includegraphics[width=0.8\linewidth]{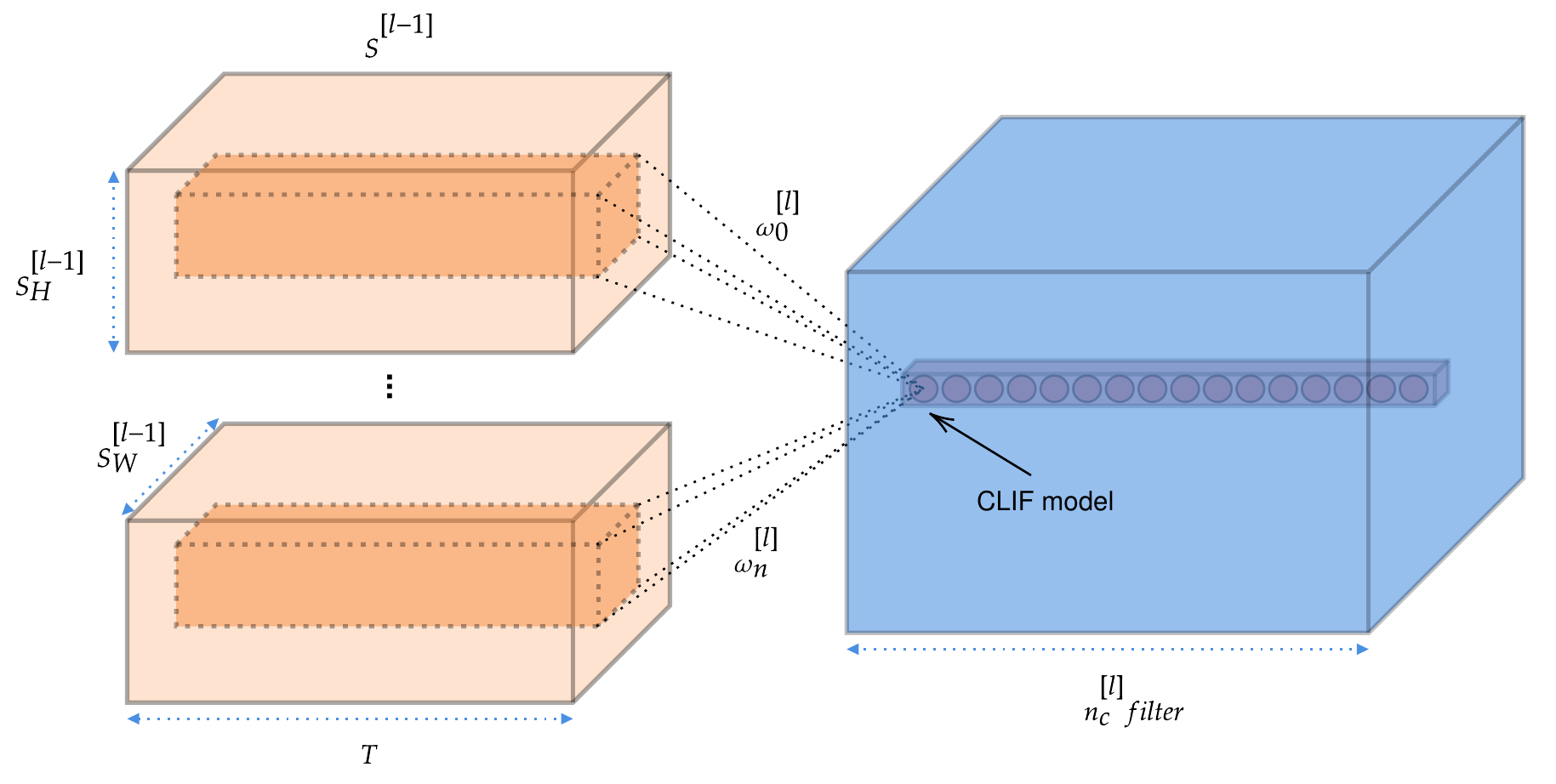}
	\caption{Spiking convolutional layer}
	\label{fig:spikeconvolution}
\end{figure}

Convolutional layers are the major building blocks used in CNNs. Through the convolution operation, the convolutional layer encodes the feature representation of the input at multiple hierarchical levels. Establishing the spiking convolutional layer is very important for building a deep SNN. The connections in convolutional layers and between the layers are similar to those in the CNN architecture. Suppose we perform a convolution operation on the output of the upper layer network, thus the input of convolutional layer can be expressed as a matrix $S^{[l-1]}$ with size $(S^{[l-1]}_{H},S^{[l-1]}_{W},S^{[l-1]}_{C},T)$. The convolutional layer passes a series of filters $\omega^{[l]}$ over our image and gets the feature map. And the filters should have the same number of channels with input $S^{[l-1]}$. The convolutional layer is summed up in Fig.\ref{fig:spikeconvolution} and the number of filters is $n^{[l]]}_{c}$. In this simulation, we add padding around the input with zero spike trains in order to maker the output size is the same as the input size (when stride = 1). Linear LIF models receive the spike train, which is the sum of elementwise multiplication of the filter and the subcube of input spike trains. This yields :

\begin{equation}
	conv(S,\omega)_{x,y} = \sum_{i=1}^{n^H} \sum_{j=1}^{n^W} \sum_{k=1}^{n^C} S_{x+i-1,x+j-1,k,T}  \omega_{i,j,k}
\end{equation}
where $\omega$ is the filter with the size of $(n^H,n^W,n^C)$. We insert the spike train into the membrane equation Eq.\ref{CLIF} and solve for the output spike train.

\subsubsection{Spiking max-pooling layer}

\begin{figure}[h!]
	\centering
	\includegraphics[width=0.8\linewidth]{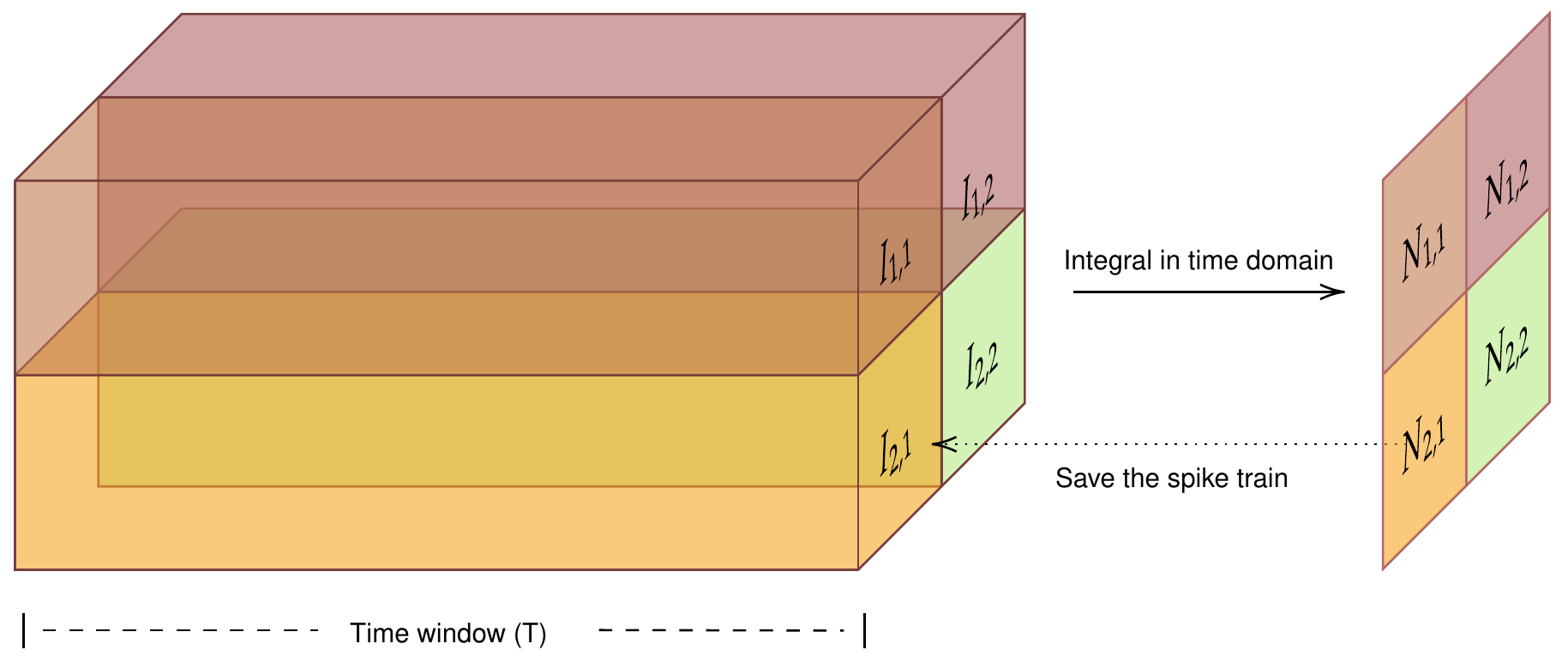}
	\caption{Spiking max-pooling layer}
	\label{fig:diagram-20210805}
\end{figure}

Most successful CNNs use max-pooling, typically added to CNNs following individual convolutional layers, to reduce computational load and overfitting. Cao \cite{cao2015spiking} used the lateral inhibition to select the winner neuron and completed the function of max-pooling layer. However, the winner neuron may not be the neuron with the largest output. Here we propose a simple method to complete the operation of the max-pooling layer. For a set of spike trains let $I_{l,h}$ denote the output of neuron. $l \in \{1,...,L\}$ and $h\in \{1,...,H\}$ represent the row and column of neuron, respectively.  Considering that we use the frequency encoding, we integrate the input spike sequences in the time domain firstly. The number of spikes $N$ of neuron $n_{l,h}$ is computed as:

\begin{equation}
	N_{l,h} = \int_{o}^{T} I_{l,h}(t)dt
\end{equation}
where $T$ is the time window. And the we adopt the neuron with the largest number of spikes in the time window as the output neuron of the max-pooling layer, shown in Fig.\ref{fig:diagram-20210805}. 

\subsubsection{Datasets and Implementation}

We evaluate our proposals on MNIST and CIFAR-10 datasets. In each group of experiments, there are two types of neural networks. One is an convolutional neural network, which performs supervised learning through backward propagation algorithm. The other is a spiking neural network built by the Linear LIF model, called LIFSNN.  The purpose of experiments is to prove the equivalence between LIFSNN and CNN, so we use the accuracy of LIFSNN compared to the output label of CNN as the evaluation criterion.

First, to prove the equivalence is established in the convolutional structure, a ConvNet with two convolution layers (Conv. 12@5 $\times$ 5 - Conv. 64@5 $\times$ 5), ReLU activations, and two max-pooling layers is trained on the MNIST dataset. The structure of networks is the same as architectures used by authors in \cite{7280696} and shown in Tab.\ref{table_CNN}. In the experiment, we select the best- performing model after the verification accuracy has converged, and directly transform it into LIFSNN. The CLIF-SNN uses frequency coding and sets the parameters of each Linear LIF neuron to be equivalent to the ReLU-AN model. 

\begin{table}[h!]
	\caption{CNN baseline model for MNIST dataset (with softmax output layer)}
	\label{table_CNN}
	\begin{tabular}{ll}
		\hline
		Layer             & Details                                                         \\ \hline
		Input layer       & 28 $\times$ 28 $\times$ 1 in $[0.0,1.0]$                  \\
		Convolution 1 & 1 $\times$ 5 $\times$ 5 kernals, ReLU,12 output maps of 28 $\times$ 28  \\
		Pooling 1         & 2 $\times$ 2 max-pooling, 12 output maps of 12 $\times$ 12 \\
		Convolution 2 & 12 $\times$ 5 $\times$ 5 kernals, ReLU,64 output maps of 12 $\times$ 12 \\
		Pooling 2         & 2 $\times$ 2 max-pooling, 64 output maps of 6 $\times$ 6   \\
		Flatten 1         & Flatten, ReLU, 3136 output maps of 1 $\times$ 1            \\
		Fully connected 1 & Fully connected, ReLU, 100 output neurons                  \\
		Fully connected 2 & Fully connected, ReLU, 10 output neurons                   \\ \hline
	\end{tabular}
\end{table}

Verified by experiment, a shallow convolutional net can achieve high performance on MNIST dataset. A more complex model should be performed to evaluate the equivalence in a deep structure. We use the AlexNet architecture \cite{krizhevsky2012imagenet} and VGG-16 \cite{simonyan2015very} architecture for the CIFAR-10 dataset. In the simulation, we did not use image pre-processing and augmentation techniques, and kept consistent with the AlexNet and VGG-16 model architecture. The AlexNet model is trained without both dropout and batch-normalization. And all the CNN in the experiment did not use bias. Because the conversion between bias and membrane conductance needs to limit the weight of the neural network, see section 4.3 for details. The equivalence between neuron models with bias has been proved in the previous chapter through formulas and simulation experiments, see section \ref{biassimulation}.

\subsubsection{Experiments for ConvNet architectures}

The network used for the MNIST dataset is trained for 100 epochs until the validation accuracy stabilizes, and achieves 98.5\% test accuracy.  For LIFSNN, we set the time window of simulation as 2 seconds and normalized values of the MNIST images to values between 0 and 10. Based on the algorithm of information coding, spike trains between 0 $Hz$ and 10 $Hz$ were generated and presented to the LIFSNN as inputs. The input trains are processed by convolutional layers and max-pooling layers, and finally are vectorized and fully connected to ten Linear LIF node as the output. We counted the number of spikes in output spike trains, used the node with the highest frequency as the output of LIFSNN.

\begin{figure}[htb!]
	\centering
	\includegraphics[angle=-90,width=0.8\linewidth]{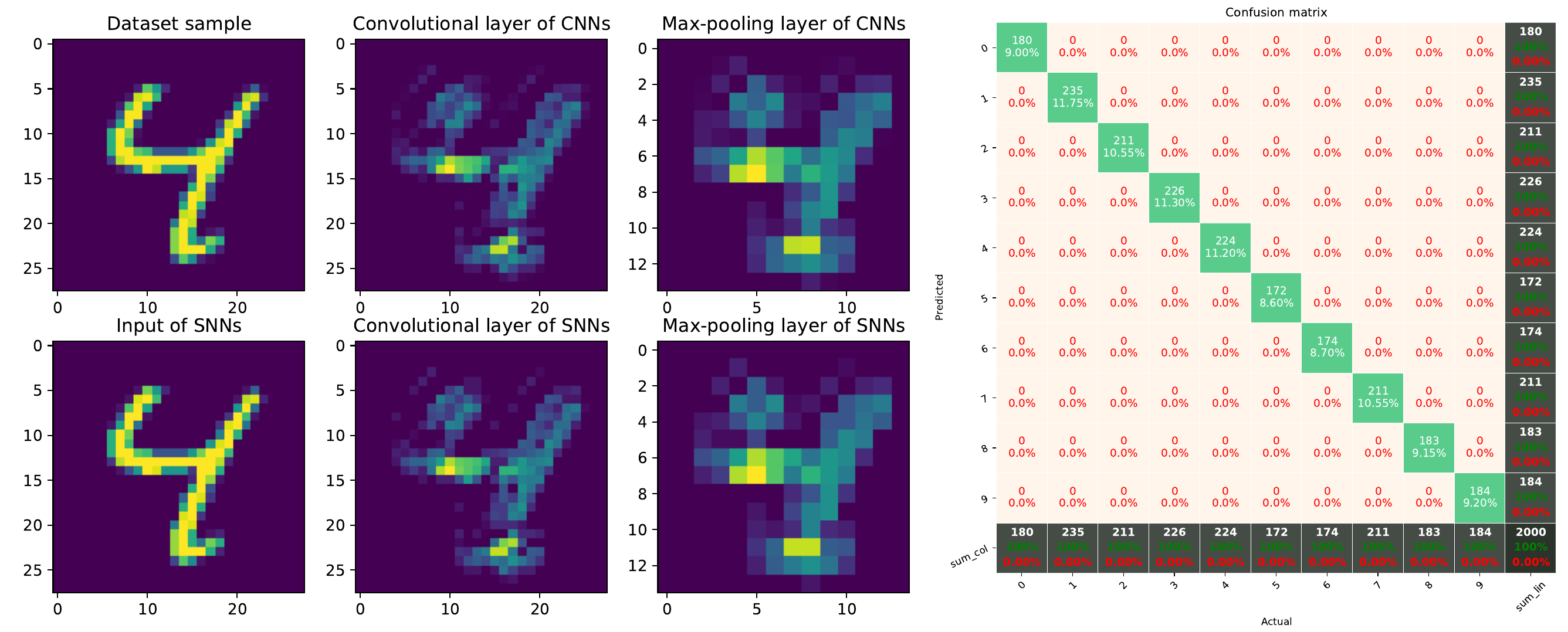}
	\caption{Performance of SNN with ConvNet architecture. The left figure shows comparison of feature maps between ReLU-based ConvNet and LIFSNN. The top figure shows the test image, feature map of convolutional layer, and feature map of max-pooling layer. The images in the bottom rows show the spikes count of output of LIFSNN. The right figure shows the confusion matrix of LIFSNN relative to the output of CNN.}
	\label{fig:mnistfig}
\end{figure}

Fig.\ref{fig:mnistfig} shows the comparison between ReLU-based ConvNet and LIFSNN, and the confusion matrix. We use the number of spikes to represent the spike trains. The comparisons of feature maps between ReLU-based ConvNet and LIFSNN are shown in the left figure in Fig.\ref{fig:mnistfig}. By comparing the upper and lower figures, we can obtain that the original images have undergone convolutional and pooling operations, which are the same as the information represented by spiking convolutional and spiking max-pooling operations after frequency encoding.  Ideally, that is, the encoding time is infinite and the sampling frequency is infinite, the image in the bottom row should be the same as the image in the top row. The right figure in Fig.\ref{fig:mnistfig} shows the confusion matrix of MNIST data, the actual labels are the outputs of CNN, and the predicted labels are the outputs converted LIFSNN. We selected 2000 sets of images from the test dataset for testing. Compared with the output of CNN, the accuracy of LIFSNN reached 100\%. Under the structure of the convolutional and pooling layers, the two neuron models can also maintain high behavioral equivalence. The experiments also proved the equivalence of the ReLU-AN model and the CLIF model in the convolutional neural network composed of convolutional and max-pooling layers. 

\subsubsection{Experiments for deep convolutional architectures}

In this subsection,  a more thorough evaluation using more complex models (e.g., VGG, AlexNet) and datasets (e.g., CIFAR that includes color images) are given. Since the main contribution of this work is establishing the mapping relationship and not in training a SOTA model. In the training of Alexnet and VGG-16 based on the CIFAR-10 dataset, we did not use data augmentation and any hyper-parameter optimization. Although the classification accuracy based on the existing training mechanism is not the best, it is already competitive.

\begin{figure}[htb!]
	\centering
	\includegraphics[width=0.8\linewidth]{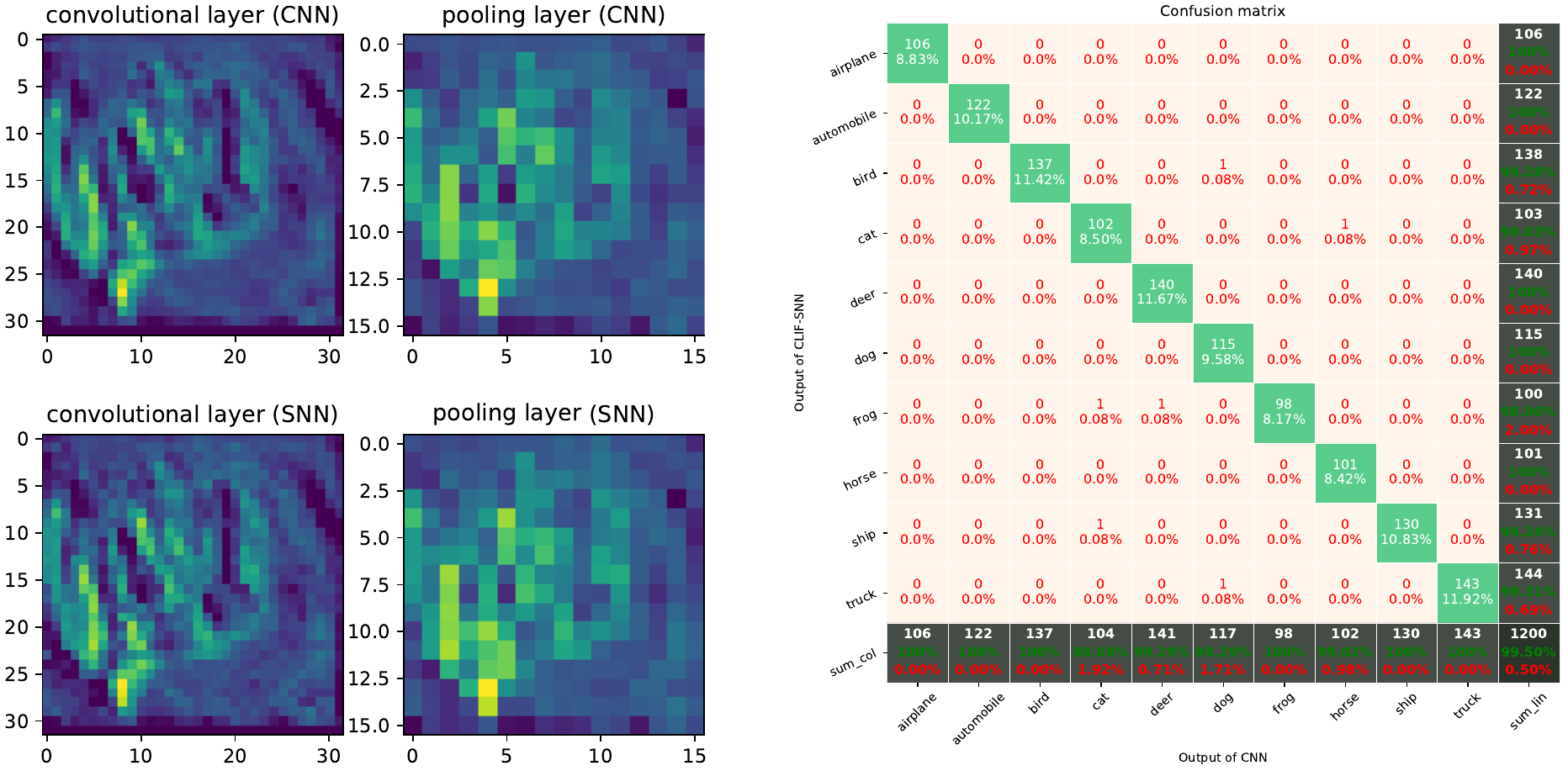}
	\caption{Performance of SNN with AlexNet architecture. The left part of figures shows the comparison of feature maps of convolutional layers and max-pooling layer between ReLU-based CNN and LIFSNN with same input and connection weights. The right part of figure shows the confusion matrix for CIFAR-10 classification.}
	\label{fig:cifarfig}
\end{figure}

The AlexNet architectures network with 5 convolutional layers, ReLU activation, $2 \times 2$ max-pooling layers after the 1st, 2nd and 5th convolutions, followed by 3 fully connected layer was trained on the CIFAR-10 dataset.  The AlexNet network for the CIFAR-10 dataset is created based on PyTorch and trained on 2 GPUs with a batchsize of 128 for 200 epochs. Classification Cross-Entropy loss and SGD with momentum 0.9 and learning rate 0.001 are used for the loss function and optimizer. We selected the best-performance model and convert the weights to the LIFSNN with the same structure. The best validation accuracies (without data augmentation) of AlexNet for the CIFAR-10 dataset we achieved were about 80.27\%. The simulation process is the same as the simulation of the MNIST dataset. Fig.\ref{fig:cifarfig} shows the comparison of the feature map and the confusion matrix. Based on the equivalence of Linear LIF model and ReLU-AN model, the outputs of LIFSNN are infinitely close to the outputs of CNN. Besides, in order to quantitatively analyze the equivalence of LIFSNN and CNN after weight conversion, we compared the classification accuracy of the two models and drew a confusion matrix. We verified 1200 samples and used the output of CNN as the label. LIFSNN achieved 99.50\% accuracy on the CIFAR-10 dataset.  

Sengupta \cite{sengupta2019going} makes effort to generate an SNN with a deep architecture and applies it to the VGG-16 network architecture. Similarly, we trained a VGG-16 structure network without the normalization and dropout layer on the CIFAR-10 dataset. The best validation accuracies we achieved were about 88.58\%.  We only replace the ReLU-AN model with the Linear LIF model. For 800 images of the test dataset, LIFSNN obtained a test accuracy rate of 99.88\% after 2000 time steps, and the accuracy rate is calculated in the same way as Sec. \ref{mnistsubsection}. Spike-normalization can also be added to the transformation process of SNN to ensure the integrity of the VGG-16 network structure to obtain higher classification accuracy.

Tab.\ref{table6} summarizes the performance of converted LIFSNN on MNIST and CIFAR-10 datasets. We list the results of some ANN-to-SNN works and compare them based on the error increment between CNN and SNN as an indicator. Error increment refers to the gap between the classification accuracies of ANN and SNN. At the same time, we also give the network structure and parameters for reference in Tab.\ref{table6}. The transformation based on model equivalence achieved the best performance. For the shallow network, we can achieve error-free transformation, and for the deep network, we can minimize the error to 0.08\%.

\begin{table}[htb!]
\tiny
\renewcommand\arraystretch{1.8}
\caption{Classification error rate on MNIST and CIFAR-10 dataset}
\label{table6}
\begin{tabular}{lllllll}
\hline
\textbf{Dataset} &
  \textbf{Architecture} &
  \textbf{Preprocess} &
  \textbf{Synap.} &
  \textbf{ANN} &
  \textbf{SNN} &
  \textbf{Error} \\ \hline
\multirow{2}{*}{\textbf{MNIST}}    & 7-layered ConvNet [ours]                              & None                  & 0.33M  & 98.5  & 98.5  & 0.0  \\
                                   & 7-layered ConvNet \cite{7280696}                & Normalization                  & 0.33M  & 99.14 & 99.12 & 0.02 \\ \hline
\multirow{6}{*}{\textbf{CIFAR-10}} & AlexNet [ours]                                        & None                  & 12.98M & 80.27 & 80.19 & 0.08 \\
 &
  8-layered ConvNet \cite{cao2015spiking} &
  Input data preprocessing &
  7.4M &
  79.12 &
 77.43 &
  1.69 \\
 &
  6-layered ConvNet \cite{rueckauer2017conversion} &
  Parameter Normalization &
  23M &
  91.91 &
  91.85 &
  1.06 \\
                                   & 8-layered Network \cite{hunsberger2016training} & None                  & -      & 83.72 & 83.54 & 0.18 \\
                                   & VGG-16 {[}ours{]}                                                 & None                  & 33M    & 88.58 & 88.46 & 0.12 \\
                                   & VGG-16 \cite{sengupta2019going}                 & Spiking Normalization & -      & 91.7  & 91.55 & 0.15 \\ \hline

\end{tabular}
\end{table}

Through the simulation of neural networks with different structures, including shallow and deep networks, we proved the equivalence of the Linear LIF model and the ReLU-AN model. And it is verified that the conversion from CNN to SNN can also be completed in convolutional structures, deep networks, and complex data sets. 
\subsection{Error Analysis}

There is still a gap between the LIF/SNN and ReLU/DNN. We believe that the main reason for the error is that the ideal simulation conditions are not achieved. Under ideal conditions, we have infinite encoding time and infinite sampling frequency. However, considering the demand for computing power, our simulations are compromised between accuracy and computing power consumption. Besides, the mapping relationship we proposed is established under the condition that multiple inputs with the same frequency. While in more general conditions, there are still errors.

Here we explore the relationship between coding time and error. We define the output of the Linear LIF model as:

\begin{equation}
	f^{\prime}=\frac{N}{T}
\end{equation}
where $N$ is the number of action potentials of spike train within the coding time, and $T$ is the coding time. Then we assume that our expected output frequency is $f$, then:

\begin{equation}
	N=\lceil f \cdot T\rceil
\end{equation}

Then the error between the expected output frequency and the true output frequency is:

\begin{equation}
	\left|f-f^{\prime}\right|=\left|f-\frac{N}{T}\right|=\left|f-\frac{[f \cdot T\rceil}{T}\right|<\frac{1}{T}
\end{equation}

\begin{figure}[htb!]
	\centering
	\includegraphics[width=0.9\linewidth]{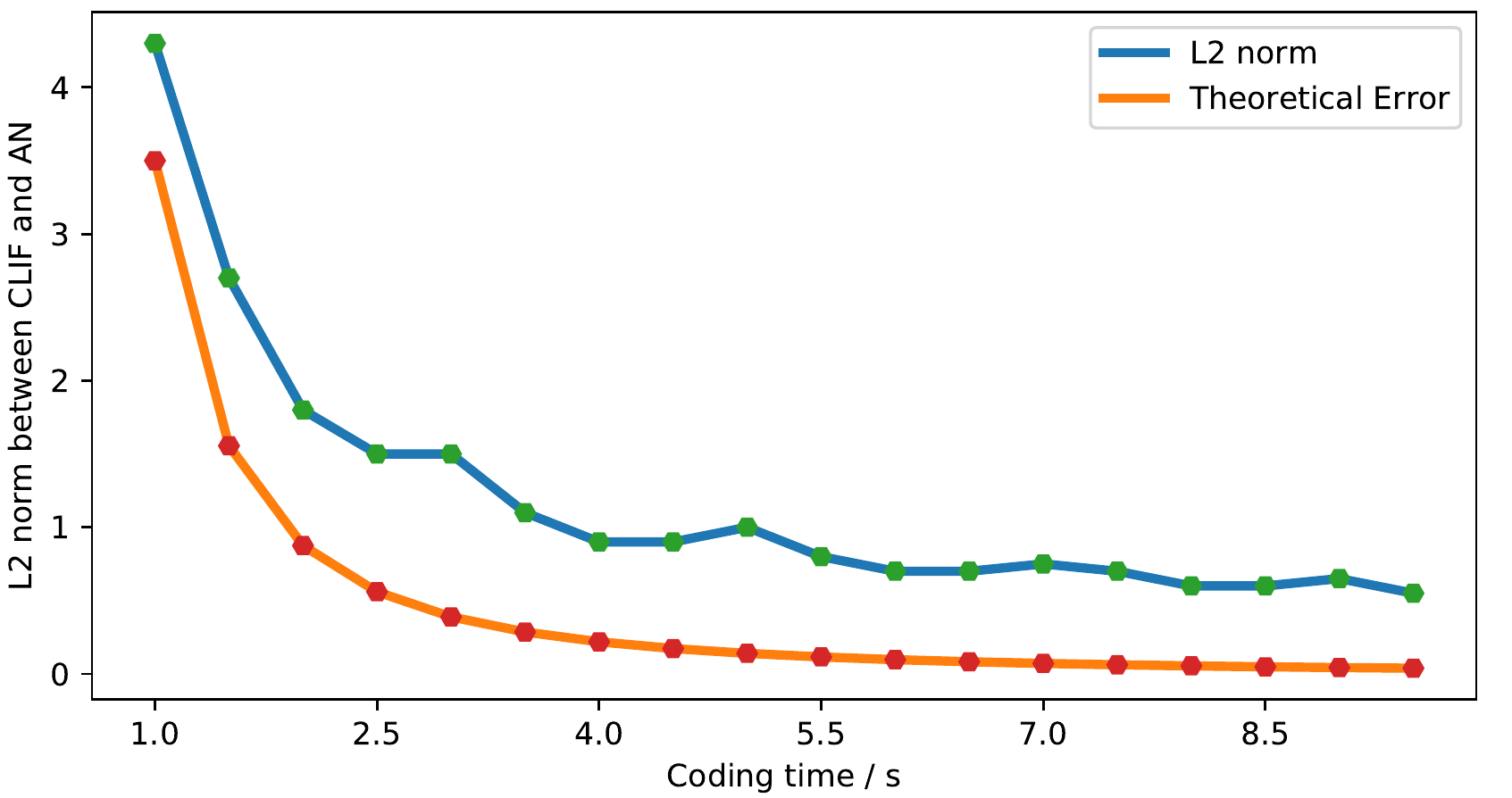}
	\caption{Comparison of theoretical error and actual error.}
	\label{fig:erroranalysis}
\end{figure}

We explore the L2 norm as the error under the same frequency input condition. Fig.\ref{fig:erroranalysis} shows the relationship between simulation error and the theoretical error of LIF-AN. We can see that the actual error is consistent with the theoretical error trend, and we can reduce the error by increasing the encoding time. When the coding time is 10s, LIFSNN achieves an error of less than 1\% in the moto/face and MNIST data sets.

\section{Conclusion and discussion}

\subsection{Brief summary}

Despite the great successes of DNN in many practical applications, there are still shortcomings to be overcome. One way to overcome them is to look for inspiration from neuroscience, where SNNs have been proposed as a biologically more plausible alternative.

This paper aims to find an equivalence between LIF/SNN and ReLU/DNN. Based on a dynamic analysis of the Linear LIF model, a parameter mapping between the biological neuron model and the artificial neuron model was established. We analyzed the equivalence of the two models from the aspects of weight, bias, and slop of activation function, and verified it both theoretically and experimentally, from a single neuron simulation to a neural network simulation. It shows that such an equivalence can be established, both the structural equivalence and behavioral equivalence, and the Linear LIF model can complete the information integration and the information processing of the linear rectification.

This mapping is helpful for the combination of an SNN with an artificial neural network and increasing the biological interpretability of an artificial neural network. It is the first step towards answering the question of how to design more causal neuron models for future neural networks. Many scholars believe that interpretability is the key to a new artificial intelligence revolution.

At the same time, the equivalence relationship is the bridge between machine intelligence and brain intelligence. Exploring new neuron models is still of great importance in areas such as unsupervised learning. As brain scientists and cognitive neuroscientists unravel the mysteries of the brain, the field of machine learning will surely benefit from it. Modern deep learning takes its inspiration from many areas, and it makes sense to understand the structure of the brain and how it works at an algorithmic level.

\subsection{Future opportunities}

The architecture of SNN is still limited to the structure of DNN. Compared with DNN, SNN only has the synaptic connection weights which can be trained, while the weights, bias and activation function (dynamic ReLU, Microsoft \cite{chen2020dynamic}) can be trained in DNN. Therefore, we expect Linear LIF model and the parameter mapping relationship can bring innovation to SNN from those aspects:

\subsubsection{A new way to convert ANN to SNN.}

With the new approach of converting pre-trained ANN to SNN, we will have a better expression of bias in SNN. Most conversion methods restrict the structure of ANN and directly map the weight. However, bias is also an important parameter in the deep learning network, and we can convert bias into membrane conductance $gl$ based on parameter mapping relationship. In this way, SNN and ANN can maintain high consistency and improve the effect of some tasks. Especially in the convolutional neural network, the connectable region of neurons is small, which is more conducive to the conversion of bias into the parameters in the Linear LIF model.

\subsubsection{Parameters training.}

All the parameters of LIF model can be trained or transformed, which is the fundamental difference from other SNN. Based on the parameters mapping relationship, we can map the trained parameters of DNN to the biological parameters of LIF model, to ensure that each node in SNN has its own unique dynamic properties. At the same time, we know the meaning of each parameter, and we can also carry out the direct training of parameters. In biology, it is also worth investigating whether other parameters of neurons, besides weights, will change.

\subsubsection{Dynamic activation function}

As the number of layers in the network increases, the number of spikes decreases. We generally adjust the spiking threshold to solve this problem. But we know that the shape of the action potential is essentially fixed, and the spiking threshold of neurons does not change. The membrane capacitance represents the ability to store ions, that is, the opening and closing of ion channels. So, when the number of pulses is low we can reduce the membrane capacitance and increase the membrane capacitance instead. In parameter mapping, it is similar to dynamic ReLU.

\appendix

\subsection{The Linear LIF model}\label{lifneuronmodel}

\subsubsection{The basic equation of Linear LIF model}

The essence of the LIF model is the parallel connection between membrane resistance $R_ M$ and membrane capacitance $C_ M$. Assuming membrane resistance $R_ M$ is voltage-independent, $R_ M = C $ is a constant. According to Ohms law, the current through the membrane resistance is $V/R_M$. Based on the definition of capacitor $C=Q/U$ (where $Q$ is the amount of charge, and $U$ is voltage), the membrane capacitive current is $I_ {C_m}=C_ M dV/dt$. According to the law of current conservation, we can get the basic equation of the LIF model:

\begin{equation}\label{lifbaseformula1}
	C_m \frac{dV(t)}{dt}+\frac{V(t) - V_0}{R_m}=I_{inj},
\end{equation}
where $C_ M$ is membrane capacitance, $R_ M$ is the membrane resistance and $I_{inj}$ is the input current.

The integral factor method is a conventional method for solving linear first-order differential equations. Based on this method, Eq. \ref{lifbaseformula1} can be further reduced to the equation of membrane potential varying with the input current, as shown in Eq. \ref{lifbaseformula2}.

\begin{equation}\label{lifbaseformula2}
	V\left(t\right)=e^{-\frac{t-t_0}{\tau_m}}\left[\int_{t_0}^{t}{\frac{I_{inj}\left(t^\prime\right)}{C_m}e^\frac{t^\prime-t_0}{\tau_m}dt^\prime}+V\left(t_0\right)\right]
\end{equation}
where $\tau_m = C_m \cdot R_m$ is the membrane time constant, $I_{inj}(t)$ is the input current, $C_m$ is th membrane capacitor, $R_m$ is the membrane resistor and $V(t_0)$ is the initial membrane potential.

\subsubsection{Membrane potential under spiking threshold}

In spiking neural networks, the most common coding method is frequency coding. That is, the pixel value is encoded into a periodic spike sequence. If there is a pulse, it is 1. Otherwise, it is 0. The encoding time (time window) is $T_w $, sampling frequency is the maximum frequency of encoding $R_{max}=1/ \Delta t$. Assuming that the pixel value to be encoded is $X_i \ in [0,1] $, the encoded spike sequence can be expressed as:

\begin{equation}\label{inj-input-1}
	I_{inj}=\omega_i\sum_{j=1}^{N}\delta\left(t-j\frac{1}{x_i}\right)
\end{equation}
where, $\delta(\ )$ is the delta function, $N=x_i \cdot T_w= \sum I_{inj}$ is the number of spikes in time windows $T_w$.

Eq. \ref{mp-change1} expresses the membrane potential with a periodic spike signal.

\begin{equation}\label{mp-change1}
	V\left(t=n/x_i\right)=\frac{\omega_i}{C_m}\cdot\left(1+e^{-\frac{T_i}{\tau_m}}+e^{-\frac{2T_i}{\tau_m}}+\cdots+e^{-\frac{\left(n-1\right)T_i}{\tau_m}}\right)
\end{equation}
when the neuron receives a spike from presynaptic, the membrane potential will accumulate. Without spike input, the membrane potential will decay exponentially. According to the summation formula of equal ratio sequence, we further simplify Eq. \ref{mp-change1} and obtain the variation equation of membrane potential.

\begin{equation}
	V\left(t=n/x_i\right)=\frac{\omega_i}{C_m}\cdot\frac{1-e^{-\frac{\left(n-1\right)T_i}{\tau_m}}\cdot e^{-\frac{T_i}{\tau_m}}}{1-e^{-\frac{T_i}{\tau_m}}}=\frac{\omega_i}{C_m}\cdot\frac{1-e^{-\frac{nT_i}{\tau_m}}}{1-e^{-\frac{T_i}{\tau_m}}}
\end{equation}

\subsubsection{Membrane potential change with spiking threshold}

The membrane potential accumulates with inputs $ I(t) $. Once the membrane potential $V(t)$ exceeds the spiking threshold $ V_{th} $, the neuron fires an action potential, and the membrane potential $V(t)$ goes back to the resting potential $ V_0 $. The LIF model is a typical nonlinear system. Three discrete equations can describe the charge, discharge, and fire of the LIF model:

\begin{equation}
	\begin{aligned}
		H(t) & = f(V(t-1), I(t)) \\
		S(t) & = \Theta(H(t) - V_{th})
	\end{aligned}
\end{equation}
where the $H(t)$ is the membrane potential before spike, $S(t)$ is the spike train and $f(V(t-1), I(t))$ is the update equation of membrane potential.

Once the membrane potential reaches the spiking threshold, an action potential will be exceeded. Then the membrane potential will be reset: 'reset to zero,' used, e.g., in \cite{diehl2015unsupervised}, reset the membrane potential to zero. 'linear reset' retains the attenuation term that exceeds the threshold:

\begin{equation}
V(t)=\left\{\begin{array}{cl}
H(t) \cdot (1 - S(t)) & \text { reset to zero } \\
H(t) \cdot (1 - S(t)) + (H(t) - V_{reset}) \cdot S(t) & \text { linear mode }
\end{array}\right.
\end{equation}

 The LIF neuron model with 'linear reset mode' is named the linear LIF model. \cite{diehl2016conversion} and \cite{10.3389/fnins.2017.00682} analyzed the difference between these two MP reset modes and chose the linear LIF model for simulation. We analyze the two models from the perspective of physics and information theory and determine the advantages of the linear LIF model. For the first reset mode, the membrane potential of the LIF model does not satisfy the law of conservation of energy. There are two parts of membrane potential attenuations: 'leaky', the attenuations as the form of conductance in the circuit which keeps the nonlinear dynamic properties. The other part is that when the action potential is exceeded, the membrane potential exceeding the spike threshold will be lost directly, resulting in energy non-conservation. From the perspective of information, the linear LIF neuron model maintains the nonlinearity of the model and retains the completion of information to the greatest extent.

\subsection{The mapping relationship between Linear LIF model and ReLU-AN model}\label{mapping-relationship}

In this chapter, we derive the parameter mapping relationship between the linear LIF model and the artificial neuron model (ReLU-AN) model. The relationship between input and output in the artificial neuron model can be expressed as:

\begin{equation}\label{ReLU}
	y_{j}=f(\sum_{i} \omega_{ji} x_{i} +b_{j})
\end{equation}
where $\omega_{ji}$ is the connerction weight between presynaptic neuron $j$ and postsynaptic neuron $i$. $b_i$ is the bias of unit $i$, and $f( \ )$ is the activation function.

Here we give the parameter mapping we established in Tab.\ref{paramsrelationship}, which will be discussed in detail later in this paper. 

\begin{table*}[htb!]
\centering

\begin{tabular}{cccc}

\hline
\multicolumn{2}{c}{\textbf{Parameter of ReLU}}                                    & \multicolumn{2}{c}{\textbf{Params of linear LIF}}      \\[7pt] \hline
\multicolumn{1}{c|}{\textbf{Symbol}} &
  \multicolumn{1}{c|}{\textbf{Description}} &
  \multicolumn{1}{c|}{\textbf{Symbol}} &
  \textbf{Description} \\[7pt] \hline
\multicolumn{1}{c|}{$\omega$} & \multicolumn{1}{c|}{Connection weight}            & \multicolumn{1}{c|}{$\omega$}        & Synaptic weight \\[7pt] \hline
\multicolumn{1}{c|}{$b$} &
  \multicolumn{1}{c|}{Bias} &
  \multicolumn{1}{c|}{$\frac{ -\sum{\omega}}{R_m C_m \cdot \ln ( 1-\sum{\omega} /(V_{th} C_m))}$} &
   \\[7pt] \hline
\multicolumn{1}{c|}{$k$}      & \multicolumn{1}{c|}{Slope of activation function} & \multicolumn{1}{c|}{$1/{V_{th} C_m} $} &                 \\[7pt] \hline
\end{tabular}
\caption{Parameter mapping between ReLU-AN model and linear LIF model}
\label{paramsrelationship}
\end{table*}

\subsubsection{Mapping of the weights}

We assume that the spiking frequency of the input signal is $f_j$ and the amplitude is 1, and then the signal can be expressed as:

\begin{equation}
	{I_{inj}}_k^l(t)=\sum_{i=1}^{n^l} \omega_{i}  \cdot \sum_{j=1}^{N^l_i}\delta(t-j\frac{1}{f_i})
\end{equation}
The $\omega_{i}$ is the synaptic weight between presynaptic neuron $i$ and post-synaptic neuron, $n^l$ represent the number of neuron in layer $l$, $j$ represents the $j_{th}$ action potential in the input spike train, $N^l_i$ is the number of action potentials, and $T_w$ is the time windows of simulation.

According to frequency coding, the number of action potentials can be solved by the product of spike frequency and coding time: $N^l_i=f_i \cdot T_w= \sum {I_{inj}}_k^l$. Compared with the weight integration process in ANNs, we integrate the input signal ${I_{inj}}_k^l(t)$ in the time window $[0,T_w]$ and obtain:

\begin{equation}
 \int_{0}^{T_w} {I_{inj}}_k^l dt = \sum_{i=1}^{n^l} \omega_{i}  \cdot \sum_{j=1}^{N^l_i} \int_{0}^{T_w}\delta(t-j\frac{1}{f_i})dt = T_w \cdot \sum_{i=1}^{n^l} \omega_{i} f_{i}
\end{equation}

\subsubsection{Mapping of the bias}\label{mapping-of-the-bias}
Due to the attenuation of membrane potential in the Linear LIF model, it is possible that even if the encoding time is long enough, neurons may not generate an action potential. If there is an action potential output within the encoding time, it must satisfy:

\begin{equation}
	V\left(N/f_i\right)\geq V_{th}
\end{equation}

Assuming that there are $N$ action potentials in the coding time $T_w$, we can get:

\begin{equation}
	\frac{\sum_{i=1}^{n^l}\omega_i}{C_m}\cdot\frac{1-e^{-\frac{NT_i}{\tau_m}}}{1-e^{-\frac{T_i}{\tau_m}}}\geq V_{th}
\end{equation}

Using $T_Wf_i \geq N$ and leads to:

\begin{equation}
	e^{-\frac{T_W}{\tau_m}}\le e^{-\frac{NT_i}{\tau_m}}\le1-\frac{C_mV_{th}\left(1-e^{-\frac{T_i}{\tau_m}}\right)}{\sum_{i=1}^{n^l}\omega_i}
\end{equation}

Simplify the formula and we can get:

\begin{equation}
	e^{-\frac{T_i}{\tau_m}}\geq1-\frac{\sum_{i=1}^{n^l}\omega_i\cdot\left(1-e^{-\frac{T_W}{\tau_m}}\right)}{C_mV_{th}}
\end{equation}

By taking logarithms on both sides of the equation, we can get:

\begin{equation}
	-\frac{T_i}{\tau_m}\geq ln\left[1-\frac{\sum_{i=1}^{n^l}\omega_i\cdot\left(1-e^{-\frac{T_W}{\tau_m}}\right)}{C_mV_{th}}\right]
\end{equation}

Transfering the period into frequency, the minimum input is given by

\begin{equation}
	{\sum_{i=1}^{n^l}\omega_if_i} \geq-\frac{\sum_{i=1}^{n^l}\omega_i}{\tau_mln\left[1-\frac{\sum_{i=1}^{n^l}\omega_i\cdot\left(1-e^{-\frac{T_W}{\tau_m}}\right)}{C_mV_{th}}\right]}
\end{equation}

Assume the encoding time meets the ideal condition, $T_ W \rightarrow \infty$. Then $e^{-T_W/\tau_m}\rightarrow 0$, the formula can be further simplified to:

\begin{equation}
	{\sum_{i=1}^{n^l}\omega_if_i} \geq-\frac{\sum_{i=1}^{n^l}\omega_i}{\tau_mln\left(1-\frac{\sum_{i=1}^{n^l}\omega_i}{C_mV_{th}}\right)}
\end{equation}

\subsubsection{Mapping of activation function}

The activation function determines the relationship between the integrated input and output. We focus on the non-negative and linear relationship of ReLU. For nonnegative, the output of the Linear LIF model is based on the number of action potentials, which is a non-negative value.

We assume that the membrane potential reaches the spiking threshold after the $n$ action potential, and the neuron generates action potentials, that is:

\begin{equation}
	V\left(t=n/x_i\right)=\frac{\sum_{i=1}^{n^l}\omega_i}{C_m}\cdot\frac{1-e^{-\frac{nT_i}{\tau_m}}}{1-e^{-\frac{T_i}{\tau_m}}}\geq V_{th}
\end{equation}

The above formula can be reduced to:

\begin{equation}
	1-\frac{V_{th}C_m\left(1-e^{-\frac{T_i}{\tau_m}}\right)}{\sum_{i=1}^{n^l}\omega_i} \geq e^{-\frac{nT_i}{\tau_m}}
\end{equation}

By taking logarithms on both sides, we can get:

\begin{equation}
	\frac{n}{\tau_mf_i}\geq-ln\left[1-\frac{V_{th}C_m\left(1-e^{-\frac{T_i}{\tau_m}}\right)}{\sum_{i=1}^{n^l}\omega_i}\right]
\end{equation}

With $f_o=f_{in}/ \lfloor n \rfloor$, the spike frequency of output sipke train can be expressed by

\begin{equation}
	\frac{f_i}{n}\le\frac{1}{-\tau_mln\left[1-\frac{V_{th}C_m\left(1-e^{-\frac{T_i}{\tau_m}}\right)}{\sum_{i=1}^{n^l}\omega_i}\right]}
\end{equation}

Using $ln(1-x) \approx x$, $e^{-x} \approx x$ and $1-e^{-\frac{T_i}{\tau_m}} \rightarrow 0$, we can get

\begin{equation}
	f_o=\frac{f_i}{n}\le\frac{1}{V_{th}C_m}\sum_{i=1}^{n^l}\omega_if_i
\end{equation}

\bibliographystyle{unsrt} 
\bibliography{reference.bib}

\end{document}